\documentclass{article} 
\usepackage{iclr2021_conference,times}

\usepackage{amsmath,amsfonts,bm}

\def\eqref#1{equation~\ref{#1}}

\def\1{\bm{1}}

\DeclareMathAlphabet{\mathsfit}{\encodingdefault}{\sfdefault}{m}{sl}
\SetMathAlphabet{\mathsfit}{bold}{\encodingdefault}{\sfdefault}{bx}{n}

\newcommand{\R}{\mathbb{R}}

\usepackage{hyperref}
\usepackage{url}
\usepackage{algorithm}
\usepackage{algorithmic}
\usepackage{multirow}
\usepackage{hhline}
\usepackage{graphicx}
\usepackage{subfigure}

\title{KDLSQ-BERT: A Quantized Bert Combining Knowledge Distillation with Learned Step Size Quantization}

\author{Jing Jin\thanks{} , Cai Liang, Tiancheng Wu, Liqin Zou, Zhiliang Gan \\
Central Software Institute, Huawei \\
\texttt{\{jinjing12,liangcai1,wutiancheng,zouliqin,ganzhiliang\}@huawei.com}
}

\iclrfinalcopy 
\begin{document}

\maketitle

\begin{abstract}
Recently, transformer-based language models such as BERT have shown tremendous performance improvement for a range of natural language processing tasks. However, these language models usually are computation expensive and memory intensive during inference. As a result, it is difficult to deploy them on resource-restricted devices. To improve the inference performance, as well as reduce the model size while maintaining the model accuracy, we propose a novel quantization method named KDLSQ-BERT that combines knowledge distillation (KD) with learned step size quantization (LSQ) for language model quantization. The main idea of our method is that the KD technique is leveraged to transfer the knowledge from a "teacher" model to a "student" model when exploiting LSQ to quantize that "student" model during the quantization training process. Extensive experiment results on GLUE benchmark and SQuAD demonstrate that our proposed KDLSQ-BERT not only performs effectively when doing different bit (e.g. 2-bit $\sim$ 8-bit) quantization, but also outperforms the existing BERT quantization methods, and even achieves comparable performance as the full-precision base-line model while obtaining 14.9x compression ratio. Our code will be public available.
\end{abstract}

\section{Introduction}
Recently, transformer-based language models such as BERT \cite{devlin2018bert} and RoBerta\cite{liu2019roberta} have achieved remarkable performance on many natural language processing tasks. However, it is difficult to deploy these models on resource-restricted devices directly since they usually contain lots of weight parameters that are computation expensive and memory intensive. To alleviate this problem, many various approaches have been widely explored to compress the model size. For instance, low-rank approximation \cite{ma2019tensorized}, \cite{lan2019albert}, weight-sharing \cite{dehghani2018universal}, \cite{lan2019albert}; knowledge distillation \cite{sanh2019distilbert},\cite{sun2019patient}, \cite{jiao2019tinybert}; pruning \cite{michel2019sixteen}, \cite{voita2019analyzing}, \cite{fan2019reducing}; dynamic network training \cite{liu2019roberta}, \cite{hou2020dynabert} and quantization \cite{zafrir2019q8bert}, \cite{shen2020q}, \cite{fan2020training}, \cite{krishnamoorthi2018quantizing}, etc.

Compared with other compression methods, quantization is widely used to compress neural network models due to two main reasons. First, quantization is an important technique enabling low-power and high-throughput DNN inference, especially it is hardware-friendly for the inference on those the resource-restricted devices such as cell-phones. Second, quantization does not change the model architecture, thus it is particularly useful for the carefully-designed network such as Transformers \cite{NIPS2017_7181}. Furthermore, 8-bit quantization is successfully applied to Transformer-based models since it can achieve a comparable performance as the full-precision baseline \cite{prato2019fully}, \cite{zafrir2019q8bert}. However, it is challenging to quantize these models within ultra low-bit (e.g., 2-bit) because of the dramatically accuracy dropping. To alleviate this issue, many complex quantization methods are proposed, mainly including mixed-precision quantization \cite{shen2020q}, \cite{zadeh2020gobo}, product quantization (PQ) \cite{fan2020training}, and ternary quantization \cite{zhang2020ternarybert}. However, different quantization methods include different features: mixed-precision quantization is not always hardware-friendly, PQ needs extra clustering operations during the training process, and ternary quantization is not a generative method that focus only on 2-bit quantization rather than on arbitrary bit quantization.

In addition to quantization, knowledge distillation (KD) \cite{hinton2015distilling} is another widely used compression method in which a compact model ("student") is trained to reproduce the behaviour of a larger model ("teacher"). For natural language processing tasks, KD has been extensively studied \cite{hu2018attention}, \cite{jiao2019tinybert}. However, instead of being used to compress models individually, KD also can be exploited by combining with other compression techniques \cite{mccarley2019pruning}, \cite{zhang2020ternarybert}, \cite{hou2020dynabert}, \cite{mao2020ladabert}.

In this work, we propose a novel BERT quantization algorithm named KDLSQ-BERT that even can achieve high accuracy via doing ultra low-bit (e.g. 2-bit) quantization. The main idea of our method is that the KD technique is leveraged to transfer the knowledge from a "teacher" BERT to a "student" one when exploiting learned step size quantization (LSQ) \cite{esser2019learned}, \cite{jain2019trained} to quantize that "student" BERT during the quantization training process. More specifically, the contributions of our work can be summarized as follows:
\begin{itemize}
  \item [1)] Inspired by \cite{jiao2019tinybert}, we use KD to transfer the knowledge from a full-precision "teacher" BERT to a quantized "student" BERT. The distillation loss in our work is calculated by the corresponding layers (such as the embedding layers and the outputs of Transformer layers) from the "teacher" BERT and the quantized "student" BERT, respectively.
  \item [2)]  Due to the excellent performance of LSQ in low-bit quantization in computer vision field \cite{jain2019trained}, we introduce such a technique to quantize the "student" BERT. To accelerate the quantization training speed and improve the model accuracy, however we present a novel method to initialize the correlated scale-factors. Furthermore, it's easy for user to fine-tune the LSQ quantization training when using such an initialization method.
  \item [3)] In fact KD can be more useful to train a "student" model only if the "teacher" model is well trained. It figures out that the distillation loss cannot work well if the "teacher" model is not well trained enough. As such, we introduce the ground truth loss built by the training data to compute the backward gradients during the training process. The extensive experiments show that the accuracy of quantized BERT can obtain a remarkable improvement if applying both distillation loss and ground truth loss.
\end{itemize}

The main work of this paper is organized as follows:  The related work regarding knowledge distillation and quantization for BERT is introduced in Section \ref{Related Work}; The main idea of our work is presented in Section \ref{Methodology}; In Section \ref{Experiments}, we do extensive experiments as well as the relevant result analysis; Finally, we summarize our work and point out the future work in Section \ref{Conclusion and Future Work}.

\section{Related Work}
\label{Related Work}
\subsection{Knowledge Distillation}
As a widely used compression technique, KD \cite{hinton2015distilling} aims at transferring the knowledge from a larger model ("teacher") to a compact model ("student") without sacrificing too much performance. Recently, many studies reveal that KD can achieve remarkable performance in a range of NLP tasks \cite{kim2016sequence}, \cite{jiao2019tinybert}. Especially, for KD on BERT compression, more comprehensive knowledge including logits, intermediate representations and attentions are adopted to train the student BERT \cite{jiao2019tinybert}, \cite{wang2020minilm}, \cite{romero2014fitnets}.

Besides, many studies show that for further model compression, KD is usually used to work with other compression techniques such as pruning \cite{mccarley2019pruning}, low-rank approximation \cite{mao2020ladabert} and dynamic networks training  \cite{hou2020dynabert}. To do so, the knowledge of the teacher BERT model can be fully leveraged, so that the student BERT can get more remarkable compression or accuracy performance improvement.

However, it should be noted that combining KD with quantization could be a promising technique for model compression as the quantized student model not only can improve the accuracy through knowledge transferring, but also can be hardware-friendly for inference. And, many related studies have been explored in convolutional neural networks (CNNs) models \cite{polino2018model}, \cite{stock2019and}, \cite{kim2019qkd}. Although TernaryBERT \cite{zhang2020ternarybert} is designed based on KD along with quantization, such a method mainly provides 2-bit and 8-bit weight quantization rather than arbitrary bit quantization. As a result, here we make our efforts to exploit KD and quantization to explore more general method for Transformer-based model quantization.

\subsection{Quantization}
\label{quantization}
Recently, quantization training has been extensively studied since the quantized models can gain a significant accuracy compensate through training process. The studies of Quantization-Aware
Training (QAT) \cite{jacob2018quantization} and Training Quantization Thresholds (TQT)\cite{esser2019learned}, \cite{jain2019trained} indicate that compared to the full-precision CNN model, the corresponding quantized model can obtain a slight accuracy drop (smaller than 1.0\%) when doing 8-bit quantization training. Besides, for the Transformers-based models, 8-bit quantization training is also successfully applied in fully quantized Transfomer (FullyQT) for machine translation \cite{prato2019fully} and Q8BERT \cite{zafrir2019q8bert}. Note that FullyQT and Q8BERT are designed based on QAT, the experimental results from  \cite{prato2019fully} and \cite{zhang2020ternarybert} show that FullyQT and Q8BERT do not work well in low-bit (e.g., 2 bit or 4-bit) quantization.

In addition, many studies has been investigated in low bit quantization for the Transformers-based models. For instance, Q-BERT \cite{shen2020q} achieves ultra-low bit quantization in terms of the analysis of fine-tuned BERT using the second order Hessian information. To avoid severe accuracy drop, mixed-precision with 3 or more bits are exploited in Q-BERT. In this case, it is usually un-friendly for inference to some hardware even though the accuracy of the quantized BERT can be enhanced to some extent. Quant-Noise (QN) \cite{fan2020training} is proposed to quantize a subset of weights in each iteration to allow unbiased gradients to flow through the model architecture. Although the high compression rate can be achieved, the quantization noise rate needs to be very carefully tuned for good performance. For TernaryBERT \cite{zhang2020ternarybert}, it is mainly used to quantize the different parts of the BERT model through 2-bit ternarization training. To avoid the accuracy drop caused by 2-bit ternarization, various distillation losses are considered to guide the quantization training process.

In this work, we extend LSQ to quantize Transformers-based models such as BERT, because the results in \cite{esser2019learned} reveal that such a quantization method is competitive in low-bit quantization for CNN models. To guarantee the accuracy of the quantized model as well as accelerate the training convergence, in our work we not only propose a novel scale-factor initialization for LSQ, but also consider various losses (including various distillation losses and ground truth loss) to guide the quantization training process.

\section{Methodology}
\label{Methodology}
\begin{figure}[htb]
\begin{center}
\includegraphics[scale=0.45]{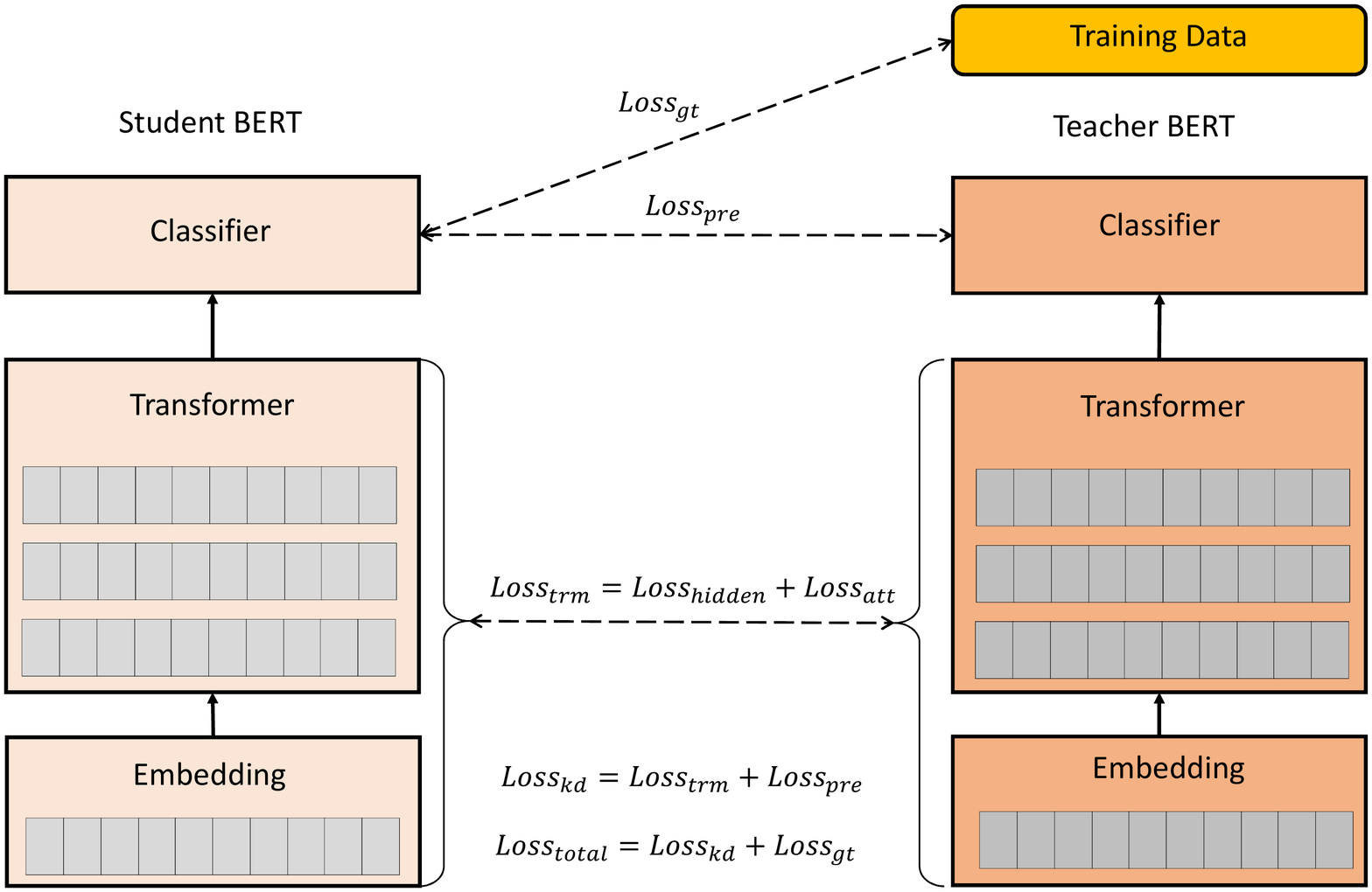}
\end{center}
\caption{The training framework of our proposed KDLSQ-BERT. In fact, the student BERT is inserted into LSQ quantization operations already after the training framework is built. The distillation loss $Loss_{kd}$ related to the knowledge of teacher BERT includes the hidden-states-based distillation loss $Loss_{hidden}$, the attention-based distillation loss $Loss_{att}$, as well as the prediction-layer-based distillation loss $Loss_{pre}$.}
\label{figure:1}
\end{figure}

In this section, we make our efforts to combine knowledge distillation with LSQ to quantize BERT. Different from the previous work \cite{zhang2020ternarybert} and \cite{jacob2018quantization}, our method has much broader capacity in different bit quantization,  meaning that it can be used not only in low-bit (e.g., 2-bit) quantization but also in high-bit (e.g., 8-bit) quantization. The main quantization training framework of our method is described in Figure \ref{figure:1}. On the one hand, the LSQ quantization operations (see \eqref{quant_and_dequant}) need to be inserted into the student model already when doing quantization training. Note that the tensors that need to be quantized include weights as well as the inputs of all linear layers and matrix multiplications. On the other hand, the total loss $Loss_{total}$ (see \eqref{total_loss}) is calculated based not only on the teacher model, but also on the ground truth from the training data set. Of course, the ground truth loss $Loss_{gt}$ (see \eqref{total_loss}) can be optional if the teacher model is well-trained. This is because the distillation loss $Loss_{kd}$ (see \eqref{kd_loss}) is enough for the student model to do quantization training if a well-trained teacher is provided. For more details on our proposed method, please see the next further discussion.

\subsection{Transformer Layer}
\label{transformer}
Most of the recent language models (e.g., BERT \cite{devlin2018bert}, XLNet \cite{yang2019xlnet}, RoBerta \cite{liu2019roberta} and TinyBert \cite{jiao2019tinybert}) are built with several Transformer layers, the main feature of which is to capture long-term dependencies between input tokens by self-attention mechanism. A standard Transformer layer includes two main sub-layers: Multi-Head Attention (MHA) and fully connected Feed-Forward Network (FFN).

For the $l$-th Transformer layer, assuming its input is $\mathbf{H}_l \in \R^{n \times d}$, where $n$ and $d$ denote the sequence length and hidden state size, respectively. Besides, suppose there are $N_H$ attention heads in each Transformer layers, and head $h$ is parameterized by $\mathbf{W}^Q_h, \mathbf{W}^K_h, \mathbf{W}^V_h, \mathbf{W}^O_h \in \R^{d \times d_h}$, where $d_h = \frac{d}{N_H}$. Then the output of head $h$ can be denoted as:
\begin{equation}\label{head_h}
  \textrm{head}_h (\mathbf{H}_l)=\textrm{Softmax}(\frac{1}{\sqrt{d}}\mathbf{H}_l \mathbf{W}^Q_h \mathbf{W}^{K \top}_h \mathbf{H}^{\top}_l)\mathbf{H}_l\mathbf{W}^V_h
\end{equation}
Let $\mathbf{W}^*=[\mathbf{W}^*_1, ...,\mathbf{W}^*_{N_H}]$, where $*$ could be anyone of $Q$, $K$ and $V$, then the output of MHA is calculated by:
\begin{equation}\label{mha}
  \textrm{MHA}_{\mathbf{W}^Q, \mathbf{W}^K, \mathbf{W}^V, \mathbf{W}^O}(\mathbf{H}_l)= \textrm{Concat}(\textrm{head}_1, ..., \textrm{head}_{N_H})\mathbf{W}^O
\end{equation}

Suppose two linear layers in FFN are parameterized by $\mathbf{W}^1 \in \R^{d \times d_{ff}}$, $\mathbf{b}^1 \in \R^{d_{ff}}$ and $\mathbf{W}^2 \in \R^{d_{ff} \times d}$, $\mathbf{b}^1 \in \R^{d}$ respectively, where $d_{ff}$ indicates the number of neurons in the intermediate layer of FFN. Meanwhile, let $\mathbf{X}_l \in \R^{n \times d}$ denote the input of FFN, the corresponding output is then computed as:
\begin{equation}\label{ffn}
  \textrm{FFN}(\mathbf{X}_l) = \textrm{GeLU}(\mathbf{X}_l\mathbf{W}^1+\mathbf{b}^1)\mathbf{W}^2+\mathbf{b}^2
\end{equation}

Using \eqref{mha} and \eqref{ffn}, the forward propagation for the $l$-th Transformer layer can be written as:
\begin{equation}\label{transformers_layer}
\begin{aligned}
     \mathbf{X}_l &= \textrm{LN}(\mathbf{H}_l+\textrm{MHA}(\mathbf{H}_l)) \\
    \mathbf{H}_{l+1} &= \textrm{LN}(\mathbf{X}_l+\textrm{FFN}(\mathbf{X}_l))
\end{aligned}
\end{equation}
where $\textrm{LN}$ is the layer normalization. Note that the input $\mathbf{H}_1$ for the first Transformer layer is from the embedding layer, so it can be denoted as:
\begin{equation}\label{embedding}
\mathbf{H}_1 = \textrm{EMB}_{\mathbf{W}^E, \mathbf{W}^S, \mathbf{W}^P}(\mathbf{z})
\end{equation}
where $\mathbf{z}$ is the input sequence, and $\mathbf{W}^E, \mathbf{W}^S$ and $\mathbf{W}^P$ are the learnable word embedding, segment embedding and position embedding, respectively.
\subsection{Learned Step Size Quantization for BERT}
\label{lsq_bert}
Typically, model quantization is useful for inference acceleration because of less computing-resource consumption of the low precision integer operations. Inspired by \cite{esser2019learned}, here we explore to quantize BERT by LSQ, the basic feature of which is that the scale-factors can be learned during the quantization training process. Specifically, the basic equations for quantization and de-quantization on LSQ are as follows:
\begin{equation}\label{quant_and_dequant}
\begin{aligned}
     q_v &= \textrm{round}(\textrm{clamp}(\frac{v}{s}, -Q_n, Q_p)) \\
      \hat{v} &= q_v \times s
\end{aligned}
\end{equation}
where $\textrm{clamp}(z, r_1, r_2)=\textrm{min}(\textrm{max}(z, r_1), r_2)$, $v$ is a real number that needs to be quantized, $q_v$ is the quantized integer number of $v$, $s$ denotes the scale-factor, $\hat{v}$ is the de-quantization result, and $Q_n$ and $Q_p$ are the negative and positive quantization levels respectively. Given a bit number $b$ for quantization, then for unsigned tensor (e.g., activations), $Q_n=0$ and $Q_p=2^b-1$; for signed tensor (e.g., weights), $Q_n=2^{b-1}-1$ and $Q_p=2^{b-1}-1$. Consequently, LSQ can be considered as a symmetric quantization.

Note that LSQ provides a way to learn $s$ in term of the training loss, then in terms of \eqref{quant_and_dequant}, the gradient to $s$ can be calculated as:
\begin{equation}\label{lsq_gradient}
\frac{\partial \hat{v}}{\partial s} = \left\{ \begin{array}{lll}
-\frac{v}{s} + \textrm{round}(\frac{v}{s}) & \textrm{if} & -Q_n < \frac{v}{s} < Q_p \\
-Q_n & \textrm{if}& \frac{v}{s} < -Q_n\\
Q_p & \textrm{if} & \frac{v}{s} > Q_p
\end{array} \right.
\end{equation}
With STE \cite{bengio2013estimating}, the gradient through the quantization function for activations can be approximated by:
\begin{equation}\label{activations_gradient}
\frac{\partial \hat{x}}{\partial x} = \left\{ \begin{array}{ll}
1 & \textrm{if} -Q_n < \frac{x}{s} < Q_p\\
0 & \textrm{otherwise}
\end{array} \right.
\end{equation}
To avoid weights becoming permanently stuck in the clipped range, however, the gradient through the quantization function for weights is computed as:
\begin{equation}\label{weights_gradient}
\frac{\partial \hat{w}}{\partial w} = 1
\end{equation}

For weight quantization,  we follow the analysis from \cite{zhang2020ternarybert} and \cite{zafrir2019q8bert} to quantize weights $\mathbf{W}^Q, \mathbf{W}^K, \mathbf{W}^V, \mathbf{W}^O$ from all transformer layers, as well as the word embedding $\mathbf{W}^E$. However, we do not implement quantization on $\mathbf{W}^S$ and $\mathbf{W}^P$ from the embedding layer, as well as the bias in linear layers because of two reasons. First, the number of these weights is negligible, thus it has less impact on the quantized model size even if quantizing them. Second, some of them contain critical information (e.g., the position information in $\mathbf{W}^P$ affects the dependency for each word) that results in significantly accuracy dropping if they are quantized. Similar to work done by \cite{zhang2020ternarybert}, softmax operation, layer normalization and the last task-specific layer are not quantized in our work. This is because quantizing these operations will lead to accuracy degradation dramatically.

With regards to the activation quantization, the inputs of all linear layers and matrix multiplication in Transformer layers are quantized. In fact, it is a symmetric quantization, so that the quantized values distribute symmetrically in both sides of 0. That is, the zero point is 0, which is benefit to accelerate inference of the quantized model.

\begin{figure}[htb]
\begin{center}
\includegraphics[scale=0.5]{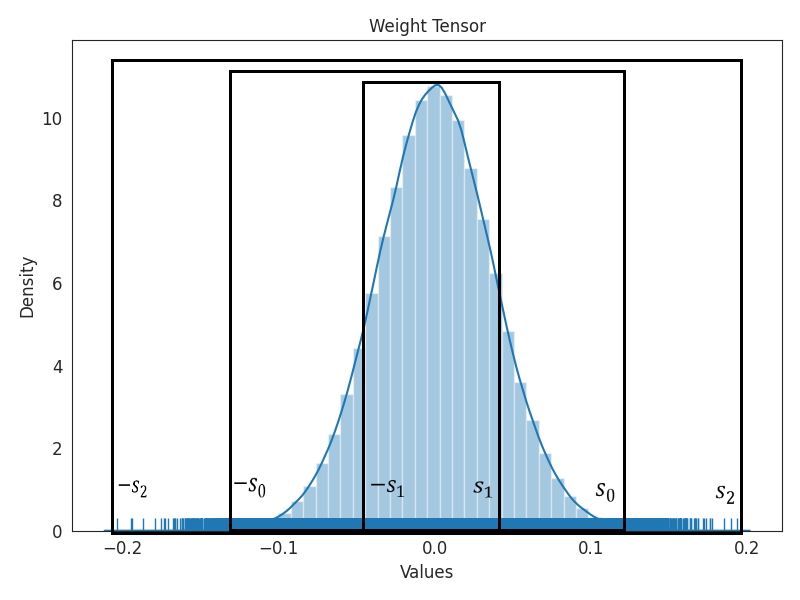}
\end{center}
\caption{The statistical histogram of a weight tensor from a well-trained TinyBERT. Compared with $s_1$ and $s_2$, $s_0$ is a much better initialized scale-factor (or truncated threshold) for quantization training since the main information of the weight tensor can be retained.}
\label{figure:2}
\end{figure}

However, we cannot expect a promising quantization result if the scale-factor $s$ for LSQ are not initialized effectively. In essence, the scale-factor can be considered as a truncated threshold for a tensor that needs to be quantized. To analyze its importance, Figure \ref{figure:2} presents the relationship between different scale-factors and the statistical histogram of a weight tensor from a well-trained TinyBERT. As illustrated in the figure, $s_0$ is a much better initialization than $s_1$ and $s_2$ since most parts of the tensor are retained by using $s_0$. In contrast, if using $s_1$, only a few parts of that tensor are retained, so that main information of that tensor will be lost. Obviously, if $s_2$ is employed, lots of boundary elements that are not in the main distribution of that tensor will be kept for quantization. In this case, such invalid information must result in accuracy dropping especially when doing low-bit quantization. Many readers may have a question that scale-factor will be updated as the training goes on, why shouldn't the scale-factor be initialized randomly? Indeed, although the scale-factor changes during the training process, an effective initialization is good to accelerate the loss convergence, as well as improve the model accuracy. Particularly, it will lead to difficult convergence for low-bit LSQ quantization training when the scale-factor is not effectively initialized.

To alleviate aforementioned issue, we propose a method to initialize the scale-factor for a given tensor that needs to be quantized. The main steps for the method are listed as follows:

\begin{algorithm}[htb]
\caption{A method to initialize scale-factor }
\label{alg:scale_factor_initialization}
\begin{algorithmic}[1]
\REQUIRE A given tensor $M$, a truncated ratio $\gamma$
\ENSURE An initialized scale-factor $s_{init}$
\STATE {$n=\textrm{nelement}(M)$, where $n$ is the number of elements in $M$.}
\STATE {$M_{sort}=\textrm{Sort}(M)$, where $M_{sort}$ is a sorted $M$ in ascending order.}
\STATE {$\textrm{index}_{min} = \textrm{round}(\frac{\gamma \times n}{2})$}
\STATE {$\textrm{index}_{max} = n - \textrm{index}_{min}$}
\IF {$\textrm{abs}(M_{sort}(\textrm{index}_{min}))\geq \textrm{abs}(M_{sort}(\textrm{index}_{max}))$}
\STATE{$s_{init} = \textrm{abs}(M_{sort}(\textrm{index}_{min}))$}
\ELSE
\STATE{$s_{init} = \textrm{abs}(M_{sort}(\textrm{index}_{max}))$}
\ENDIF
\RETURN {$s_{init}$ }
\end{algorithmic}
\end{algorithm}

From algorithm \ref{alg:scale_factor_initialization}, it shows that the truncated ratio $\gamma$ (where $\gamma \ll 1$) determines how many elements in $M$ need to be truncated. With $\gamma$ and the sorted $M_{sort}$, $s_{init}$ is initialized to a value that can be regarded as a threshold for truncation. More specifically, the elements that are greater than $s_{init}$ need to be truanted to $Q_p$, the elements that are smaller than $-s_{init}$ need to be truanted to $-Q_n$, so the most of elements in a range of $(-s_{init}, s_{init})$ are not changed. That's, since the elements in $M$ are truanted in terms of $s_{init}$, the majority information of $M$ can be retained for quantization.

As analyzed previously, both weights and activations need to be quantized by LSQ. For weights, we make use of the corresponding weight values from the full-precision BERT to do scale-factor initialization. Of course, the full-precision BERT should be well trained and be set as a pre-trained model for LSQ quantization training. For activations, however, it is a little bit different as activations cannot be obtained from the full-precision BERT directly. Therefore, we leverage a batch of training data set (which is selected randomly) to do inference for the full-precision BERT, such that the correlated activations can be obtained. After the activations are calculated, the related scale-factors can be initialized according to Algorithm \ref{alg:scale_factor_initialization}.

\subsection{Distillation-Aware Quantization}
\label{distill_aware_quantization}
For quantization training, although the model accuracy loss introduced by quantization can be compensated through training process, it may be less effective when doing low-bit quantization. To address such a problem, thus we take advantage of KD to further improve the accuracy performance of the quantized BERT. Following the teacher-student KD framework, we set the quantized BERT and the well-trained full-precision BERT to be the student model and the teacher model, respectively. For the quantized BERT, the related LSQ quantization operations are inserted already, and it can learn to recover the behaviours of the teacher model via the KD technique.

Inspired by \cite{zhang2020ternarybert} and \cite{jiao2019tinybert}, the distillation loss for the Transformer layers mainly relies on the outputs of all Transformer layers, the outputs of embedding layer, as well as the attention scores of all heads from all Transformer layers. More specifically, considering the outputs of all Transformer layers and the outputs of embedding layer, the hidden-states-based distillation loss $Loss_{hidden}$ can be calculated as follows:
\begin{equation}\label{hidden_states_loss}
Loss_{hidden}=\sum_{l=1}^{L+1}\textrm{MSE}(\mathbf{H}_l^S, \mathbf{H}_l^T)
\end{equation}
where MSE indicates the mean squared error, $L$ is the number of Transformer layers for BERT, $\mathbf{H}_l^S$ ($\mathbf{H}_l^T$) is the inputs of the $l$-th Transformer layer from the student model (teacher model). However, note that $\mathbf{H}_{L+1}^S$ ($\mathbf{H}_{L+1}^T$) denotes the outputs of the last Transformer layer from the student model (teacher model). For the attention-based distillation loss $Loss_{att}$ \cite{clark2019does} related to the attention scores of all heads from all Transformer layers, it can be computed by:
\begin{equation}\label{attention_loss}
Loss_{att}=\sum_{l=1}^{L}\textrm{MSE}(\mathbf{A}_l^S, \mathbf{A}_l^T)
\end{equation}

Consequently, the distillation loss $Loss_{trm}$ for all Transformer layers is computed by:
\begin{equation}\label{attention_loss}
Loss_{trm}=Loss_{hidden}+Loss_{att}
\end{equation}
In addition to Transformer layers, the knowledge from the prediction layer affects the accuracy performance significantly. Therefore, we also distill that knowledge, such that the logits $\mathbf{P}^S$ from the student model can learn to fit the corresponding $\mathbf{P}^T$ from the teacher model by soft cross-entropy (SCE) loss:
\begin{equation}\label{attention_loss}
Loss_{pre}=\textrm{SCE}(\mathbf{P}^S, \mathbf{P}^T)
\end{equation}

Consequently, the total distillation loss $Loss_{kd}$ established between the student model and the teacher model can be computed by:
\begin{equation}\label{kd_loss}
Loss_{kd}= Loss_{pre}+Loss_{trm}
\end{equation}

However, both theoretical and empirical discoveries show that $Loss_{kd}$ can work well for training only if the teacher model is well-trained. That's, a bad teacher model results in a bad accuracy for the student model. To avoid the unexpected result caused by a bad teacher model, we introduce the ground truth loss $Loss_{gt}$ that needs to be calculated according to the label truth of the training data set. In doing so, the quantization training loss for the student model is not only related to the teacher model, but also related to the training data set. Thus, the total training loss can be computed by:
\begin{equation}\label{total_loss}
Loss_{total}=Loss_{kd}+Loss_{gt}
\end{equation}

According to the analysis above, the whole procedure of our distillation-aware LSQ quantization can be summarized as follows:
\begin{algorithm}[htb]
\caption{KDLSQ-BERT: A novel distillation-aware quantization based on LSQ}
\label{alg:distillation_aware_quantization}
\begin{algorithmic}[1]
\REQUIRE Training data set, A well-trained full-precision BERT.
\ENSURE A quantized student BERT
\STATE {Initialize the teacher model and student model by the well-trained full-precision BERT.}
\STATE {Initialize the scale-factors for the weights in student model in terms of Algorithm \ref{alg:scale_factor_initialization}.}
\STATE {Get activations by doing inference of the student model with a randomly selected batch of training data set.}
\STATE {Initialize the scale-factors for the above activations (including the inputs of linear layers and the inputs of the matrix multiplications) in terms of Algorithm \ref{alg:scale_factor_initialization}.}
\STATE {Insert the LSQ quantization operations into the student model according to \eqref{quant_and_dequant}.}
\STATE {Initialize the learning rate $\eta$ and $epochs$.}
\FOR {$i=1, ..., epoches$}
\FOR {$iter=1, ..., max_{batch}$ }
\STATE {Get the $iter$-th batch of training data set.}
\STATE {Compute the loss according to \eqref{total_loss}.}
\STATE {Compute the correlated gradients according to \eqref{lsq_gradient}, \eqref{activations_gradient}, and \eqref{weights_gradient}.}
\STATE {Update the weights and the scale-factors for the student model.}
\STATE {Update the learning rate $\eta$.}
\ENDFOR
\ENDFOR
\end{algorithmic}
\end{algorithm}

\section{Experimental Analysis}
\label{Experiments}
In this section, we main evaluate the performance of our proposed KDLSQ-BERT on both the GLUE benchmark \cite{wang2018glue} and SQuAD \cite{rajpurkar2016squad}. The specific information for those data sets are listed as follows:

\textbf{The GLUE benchmark:} The GLUE benchmark is a collection of diverse natural language understanding tasks, including textual entailment (RTE), natural language inference (MNLI, QNLI), similarity and paraphrashe (MRPC, QQP, and STS-B), sentiment Analysis (SST-2) and linguistic acceptability (CoLA). For the performance evaluation metrics \cite{chicco2020advantages}, we follow the work of TinyBERT\cite{jiao2019tinybert} by setting Matthews Correlation for CoLA, F1 for MRPC and QQP, Spearman Correlation for STS-B, and accuracy for the other tasks.

\textbf{SQuAD:} SQuAD v1.1 is a machine reading comprehension task. Given a question-passage pair, the task is to extract the answer span from the passage. SQuAD v2.0 is a updated version in which the question might be unanswerable. For the performance evaluation metric \cite{derczynski2016complementarity}, also we follow the work of TinyBERT\cite{jiao2019tinybert} by considering F1 for these two data sets.

In addition, the code for our proposed method is modified from the huggingface pytorch-transformer library\footnote{\url{https://github.com/huggingface/transformers}}, Q8BERT \footnote{\url{https://github.com/NervanaSystems/nlp-architect.git}} and TinyBERT\footnote{\url{https://github.com/huawei-noah/Pretrained-Language-Model/tree/master/TinyBERT}}. To effectively investigate the performance of our method, we set Q8BERT \cite{zafrir2019q8bert} and TernaryBERT \cite{zhang2020ternarybert} as the comparison methods. For Q8BERT, we set arbitrary bit (2-bit $\sim$ 8-bit) weight quantization as well as the fixed 8-bit activation quantization for fair comparison. For TernaryBERT, we consider 2-bit and 8-bit weight quantization for comparison according to the discussion did by \cite{zhang2020ternarybert}. However, we do not use Q-BERT \cite{shen2020q} as a comparison algorithm since it is not open-source. Here we abbreviate the quantization bit for weights of Transformer layers, word embedding and activations as "W-E-A (\#bit)". If not specified, "W-E-A (\#bit)" denotes the same meaning for our next experimental analysis. The used BERT models in our experiments include TinyBERT and BERT-base, and the specific information of these models are listed in Table \ref{tab:1}.

\begin{table}[htb]
\caption{The basic information for the used BERT models}
\label{tab:1}
\begin{center}
\scalebox{1.0} {
\begin{tabular}{l|cccc}
\hline \hline  \multirow{2}*{Model} & Transformer & Hidden & Feed-Forward & Model \\
                             & Layers & Size & Size & Size(MB) \\
\hline
\hline
BERT-base & 12 & 768 & 3072 & 418 ($\times 1$)\\
TinyBERT  & 6 & 768 & 3072 & 258 ($\times 1.6$)\\
\hline \hline
\end{tabular}}
\end{center}
\end{table}

For the truncated ratio $\gamma$ which is the input of Algorithm \ref{alg:scale_factor_initialization}, we set it to be $0.05$ , meaning that a tensor that needs to be quantized can be retained $95.0\%$ information for quantization. However, the initial learning rate for weight scale-factors is different from that for activation scale-factors. For weight scale-factors, we set the corresponding initial learning rate to be $1.0\times 10^{-3}$ since a well-trained pre-trained model can provide effective weight parameters for scale-factor initialization. In doing so, the small initial learning rate can bring small update size, so that the weight scale-factors can be updated slightly during the quantization training process. Differently, as the activations are estimated by a batch size of training data, such an inaccurate estimation requires that the initial learning rate of activation scale-factors should be greater than that of weight scale-factors. Therefore, we set the initial learning rate for activation scale-factors to be $2.0\times 10^{-2}$, so that the corresponding scale-factors can be updated effectively. Besides, the learning rate is set to decay linearly to 0 during the quantization training, and the number of training epoch is set to be 3.0.

\subsection{Experimental Comparison with Existing BERT Quantization Methods}
\label{experiment_comparison}
For the GLUE benchmark, we conduct the experiments with the batch size of 16 for CoLA and 32 for other tasks. The learning rate is initialized to $2 \times 10^{-5}$ and decays linearly to 0 during 3.0 training epochs. The maximum sequence length is 64 for single-sentence tasks CoLA and SST-2, and 128 for the rest sentence-pair tasks. Besides, the dropout rate for hidden representations and the attention probabilities is set to 0.1. Note that for the activation quantization, we consider all the methods by setting a fixed 8-bit quantization for the fair comparison.

\begin{table}[htb]
\caption{Experimental Comparison on the GLUE benchmark. }
\label{tab:2}
\begin{center}
\scalebox{0.60} {
\begin{tabular}{llcc|ccccccccc}
\hline \hline & & W-E-A & Size & \multirow{2}{*}{MNLI} & \multirow{2}{*}{CoLA} & \multirow{2}{*}{MRPC} & \multirow{2}{*}{QNLI} & \multirow{2}{*}{QQP} & \multirow{2}{*}{RTE} & \multirow{2}{*}{SST-2} & \multirow{2}{*}{STS-B} & \multirow{2}{*}{$Average$}  \\  & & (\#bit) & (MB) & &  & &  &  & & & & \\ \hline
\hline Full-precision & BERT-base  & 32-32-32 & 418 ($\times 1.0$) & 84.463 & 58.081 & 90.625 & 91.964 & 87.762 & 71.119 & 93.119 & 89.823 & 83.369 \\
\hline \multirow{3}{*}{2-bit} & Q8BERT & 2-2-8 & 28 ($\times 14.9$) & 38.268 & 0.000 & 81.223 & 53.139 & 10.856 & 52.708 & 50.917 & 7.441 & 36.819 \\
& TernaryBERT & 2-2-8 & 28 ($\times 14.9$) & 83.780 & \bf 49.987 & \bf 90.444 & 91.159 & 87.007 & \bf 67.870 & 92.087 & 87.732 & 81.258\\
& KDLSQ-BERT (ours) & 2-2-8 & 28 ($\times 14.9$) & \bf 84.564 & 49.418 & 90.290 & \bf 91.452 & \bf 88.071 & 67.509 & \bf 92.775 & \bf 88.117 & \bf 81.524 \\
\hline \multirow{2}{*}{4-bit} & Q8BERT & 4-4-8 & 54 ($\times 7.7$) & 70.321 & 6.577 & 75.835 & 80.029 & 80.682 & 55.235 & 82.569 & 72.358 & 65.451\\
& KDLSQ-BERT (ours) & 4-4-8 &  54 ($\times 7.7$) & \bf 85.329 & \bf 60.646 & \bf 91.349 & \bf 92.165 & \bf 88.503 & \bf 72.729 & \bf 93.349 & \bf 89.697 & \bf 84.221\\
\hline \multirow{2}{*}{6-bit} & Q8BERT & 6-6-8 & 80 ($\times 5.2$) & 82.965 & 48.518 & 89.456 & 89.438 & 87.702 & 68.953 & 91.743 & 88.163 & 80.867\\
& KDLSQ-BERT (ours) & 6-6-8 & 80 ($\times 5.2$) & \bf 85.329 & \bf 61.482 & \bf 91.003 & \bf 92.367 & \bf 88.463 & \bf 71.480 & \bf 93.807 & \bf 90.000 & \bf 84.241\\
\hline \multirow{3}{*}{8-bit} & Q8BERT & 8-8-8 & 106 ($\times 3.9$) &  84.401 & 59.937 & 90.941 & 91.122 & 87.832 & 72.202 & 93.349 & 89.569 & 83.669\\
& TernaryBERT & 8-8-8 & 106 ($\times 3.9$)&  84.422 & 57.833 & 90.744 & 91.928 & 87.433 & \bf 72.563 & 93.463 & \bf 89.945 & 83.541 \\
& KDLSQ-BERT (ours) & 8-8-8 & 106 ($\times 3.9$) &  \bf 85.308 & \bf 60.774 & \bf 91.826 & \bf 92.147 & \bf 88.439 & 71.841 & \bf 93.693 & 89.944 & \bf 84.247
\\
\hline
\hline  Full-precision & TinyBERT  & 32-32-32 & 258 ($\times 1.6$) & 84.768 & 54.175 & 91.319 & 90.793 & 87.966 & 71.841 & 90.252 & 89.792 & 82.613\\
\hline \multirow{3}{*}{2-bit} & Q8BERT & 2-2-8 & 18 ($\times 23.2$) & 39.277 & 0.000 & 81.223 & 53.890 & 44.831 & 52.708 & 51.376 & 8.039 & 41.418\\
& TernaryBERT & 2-2-8 & 18 ($\times 23.2$) & 83.719 & 48.258 & 90.625 & 89.969 & 86.759 & \bf 66.787 & 92.201 & 87.425 & 80.718 \\
& KDLSQ-BERT (ours) & 2-2-8 & 18 ($\times 23.2$) & \bf 83.902 & \bf 49.047 & \bf 91.100 & \bf 90.518 & \bf 88.038 & 64.260 & \bf 92.661 & \bf 87.539 & \bf 80.883\\
\hline \multirow{2}{*}{4-bit} & Q8BERT & 4-4-8 & 34 ($\times 12.3$)  & 78.451 & 12.517 & 81.563 & 83.031 & 85.279 & 59.206 & 87.615 & 79.696 & 70.920\\
& KDLSQ-BERT (ours) & 4-4-8 & 34 ($\times 12.3$) & \bf 84.962 & \bf 54.415 & \bf 91.192 & \bf 91.671 & \bf 88.300 & \bf 72.202 & \bf 92.890 & \bf 89.825 & \bf 83.182\\
\hline \multirow{2}{*}{6-bit} & Q8BERT & 6-6-8 & 50 ($\times 8.3$) & 83.719 &	51.903 & \bf 91.724 & 90.042 & 87.620 &  72.563 & 91.514 & 89.223 & 82.288\\
& KDLSQ-BERT (ours) & 6-6-8 & 50 ($\times 8.3$) & \bf 85.013 & \bf 54.293 & 91.035 & \bf 91.580 & \bf 88.333 & \bf 73.285 & \bf 93.006 & \bf 89.895 & \bf 83.305\\
\hline \multirow{3}{*}{8-bit} & Q8BERT & 8-8-8 & 65 ($\times 6.4$) & 83.871 &	54.015 & 90.662 &  90.500 &  87.890 & \bf 74.007 & 	92.317 & 89.802 & 82.883\\
& TernaryBERT & 8-8-8 & 65 ($\times 6.4$) & 84.697 & 54.177 & 91.130 & 90.902 & 87.856 & 72.202 & 92.890 & \bf 90.034 & 82.986\\
& KDLSQ-BERT (ours) & 8-8-8 & 65 ($\times 6.4$) & \bf 85.013 & 	\bf 54.793 & \bf 91.130 & \bf 91.580 & \bf 88.315 & 73.646 & 92.890 & 89.958 & \bf 83.416 \\
\hline
\hline
\end{tabular}}
\end{center}
\end{table}

As listed in Table \ref{tab:2}, the results demonstrate that for the same bit quantization, our proposed KDLSQ-BERT significantly outperforms the existing BERT quantization methods on the GLUE tasks. More specifically, for 2-bit weight quantization, KDLSQ-BERT achieves almost closely accuracy performance as the full-precision base-line model; for 4-bit $\sim$ 8-bit weight quantization, KDLSQ-BERT achieves much better accuracy than the full-precision base-line model on all GLUE tasks. Particularly, note that for 2-bit weight quantization of TinyBERT, KDLSQ-BERT achieves only about 1.5\% accuracy drop on average while obtaining 23.2x compression ratio compared with BERT-base.

For SQuAD related to question-answering tasks, we conduct the experiments with the batch size of 16 and the maximum sequence length of 384. The learning rate is initialized to $2 \times 10^{-5}$ and decays linearly to 0 during 3.0 training epochs. The dropout rate for hidden representations and the
attention probabilities is set to be 0.1. However, we apply a fixed 8-bit for activation quantization in our experiments so as to make a fair comparison. The specific measure results are presented in Table \ref{tab:3}.
\begin{table}[htb]
\caption{Experimental Comparison on SQuAD.}
\label{tab:3}
\begin{center}
\scalebox{0.68} {
\begin{tabular}{llcc|ccc}
\hline \hline & & W-E-A & Mode & \multirow{2}{*}{SQuAD 1.1}  & \multirow{2}{*}{SQuAD 2.0} & \multirow{2}{*}{$Average$}\\  &  & (\#bit) & (MB) & & & \\ \hline
\hline Full-precision & BERT (base)  & 32-32-32 & 418 ($\times 1.0$) & 88.696 & 77.725 & 83.210\\ \hline
\multirow{3}{*}{2-bit} & Q8BERT & 2-2-8 & 28 ($\times 14.9$) & 3.411 & 50.072 & 26.741\\
& TernaryBERT & 2-2-8 & 28 ($\times 14.9$) & 87.672 & 75.863 & 81.767 \\
& KDLSQ-BERT (ours) & 2-2-8 & 28 ($\times 14.9$) & \bf 88.447 & \bf 78.400 & \bf 83.423 \\
\hline
\multirow{2}{*}{4-bit} & Q8BERT & 4-4-8 & 54 ($\times 7.7$) & 23.817 & 50.080 & 36.948\\
& KDLSQ-BERT (ours) & 4-4-8 & 54 ($\times 7.7$) & \bf 89.207 & \bf 78.965 & \bf 84.086\\
\hline
\multirow{2}{*}{6-bit} & Q8BERT & 6-6-8 & 80 ($\times 5.2$) & 85.481 & 72.818 & 79.149\\
& KDLSQ-BERT (ours) & 6-6-8 & 80 ($\times 5.2$) & \bf 89.280 & \bf 78.905 & \bf 84.092\\
\hline
\multirow{3}{*}{8-bit} & Q8BERT & 8-8-8 & 106 ($\times 3.9$) & 86.291 & 75.486 & 80.889\\
& TernaryBERT & 8-8-8 & 106 ($\times 3.9$) & 88.721 & 77.238 & 82.979\\
& KDLSQ-BERT (ours) & 8-8-8 & 106 ($\times 3.9$) & \bf 89.218 & \bf 79.330 & \bf 84.274\\
\hline
\hline Full-precision & TinyBERT  & 32-32-32 & 258 ($\times 1.6$) & 87.527 & 77.730 & 82.629\\ \hline
\multirow{3}{*}{2-bit} & Q8BERT & 2-2-8 & 18 ($\times 23.2$) & 3.839 & 50.071 & 26.955\\
& TernaryBERT & 2-2-8 & 18 ($\times 23.2$) & 86.003 & 76.332 & 81.168\\
& KDLSQ-BERT (ours) & 2-2-8 & 18 ($\times 23.2$) & \bf 86.053 & \bf 76.889 & \bf 81.762\\
\hline
\multirow{2}{*}{4-bit} & Q8BERT  & 4-4-8 & 34 ($\times 12.3$) & 77.320 & 50.071 & 63.696\\
& KDLSQ-BERT (ours) & 4-4-8 & 34 ($\times 12.3$) & \bf 87.887 & \bf 77.981 & \bf 82.934\\
\hline
\multirow{2}{*}{6-bit} & Q8BERT  & 6-6-8 & 50 ($\times 8.3$) & 85.596 & 75.304 & 80.450\\
& KDLSQ-BERT (ours) & 6-6-8 & 50 ($\times 8.3$) & \bf 87.944 & \bf 78.279 & \bf 83.111 \\
\hline
\multirow{3}{*}{8-bit} & Q8BERT & 8-8-8 & 65 ($\times 6.4$) & 86.375 & 75.841 & 81.108\\
& TernaryBERT & 8-8-8 & 65 ($\times 6.4$) & 87.630 & 77.464 & 82.547\\
& KDLSQ-BERT (ours) & 8-8-8 & 65 ($\times 6.4$) & \bf 87.948 & \bf 78.302 & \bf 83.125\\
\hline \hline
\end{tabular}}
\end{center}
\end{table}

From the table, it shows that for SQuAD v1.1 and SQuAD v2.0, KDLSQ-BERT significantly outperforms Q8BERT and TernaryBERT in the same bit quantization, and meanwhile it achieves even comparable accuracy as the full-precision base-line model. More specially, for 2-bit weight quantization of TinyBERT, KDLSQ-BERT achieves almost the same performance as the full-precision base-line model while being 23.2x compression ratio compared with BERT-base. Additionally, for other bit weight quantization of TinyBERT, KDLSQ-BERT performs better than not only the existing BERT quantization methods, but also the corresponding full-precision base-line model.

\subsection{Effects on Scale-factor Initialization}
\label{effect_scale_initialization}
As methioned in Section \ref{lsq_bert}, it has a significant impacts on training convergence and model accuracy when using different scale-factor initialization during KDLSQ-BERT training. However, because of the space constrains, here we choose the classic \textbf{"mnli"} from the GULE benchmark tasks and the classic \textbf{"SQuAD 1.1"} from the SQuAD tasks to do the experimental analysis. Here, we adopt two different scale-factor initialization for comparison, one is Algorithm \ref{alg:scale_factor_initialization} (indicated "ours"), and the other one is the expert experience value for scale-factor initialization (indicated "experience") . For the expert experience value, we set 4.0 to do initialization for weight scale-factors and set 16.0 to do initialization for activation scale-factors according to the extensive experiment results. Indeed, it is really not easy to find an effective expert experience value to do scale-factor initialization because of lots of experiment exploration. The correlated losses that need to do deeply analysis are presented in \eqref{total_loss}, mainly including $Loss_{total}$, $Loss_{kd}$, and $Loss_{gt}$ .

\begin{figure}[htb]
\begin{center}
\subfigure[$Loss_{total}$: 2-bit]{
\begin{minipage}[t]{0.25\linewidth}
\centering
\includegraphics[width=1.45in]{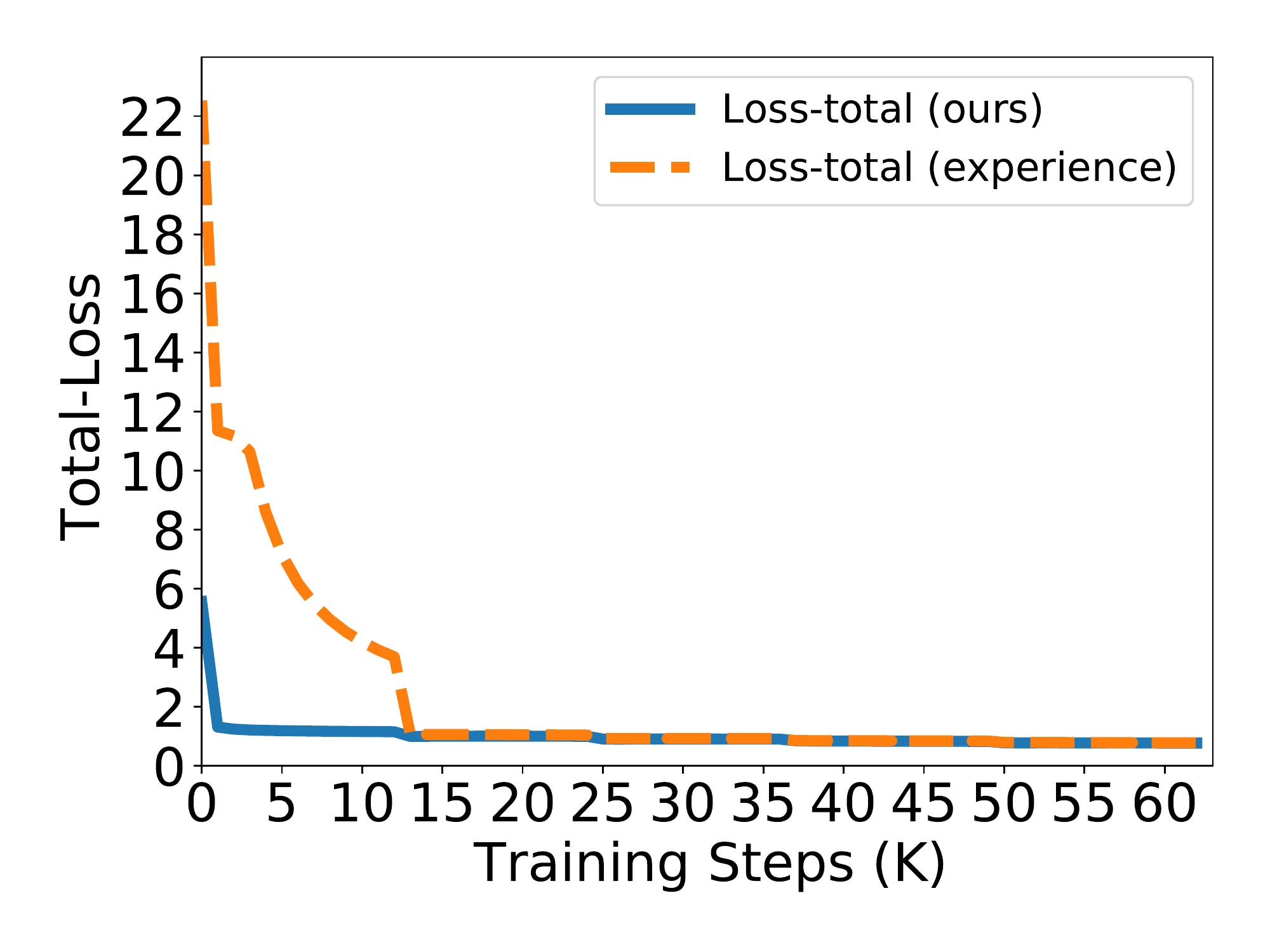}
\end{minipage}%
}%
\subfigure[$Loss_{kd}$: 2-bit]{
\begin{minipage}[t]{0.25\linewidth}
\centering
\includegraphics[width=1.45in]{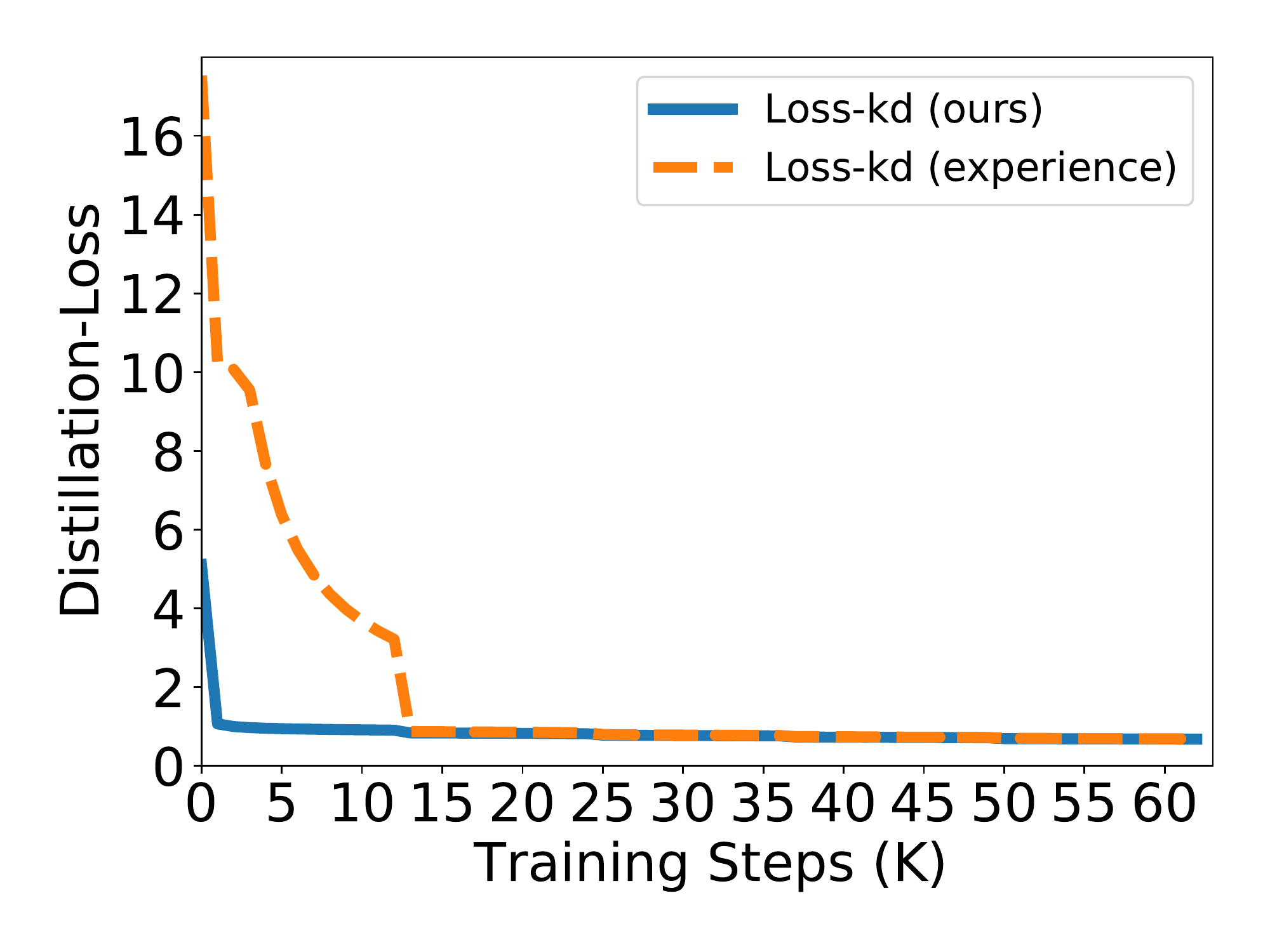}
\end{minipage}%
}%
\subfigure[$Loss_{gt}$ 2-bit]{
\begin{minipage}[t]{0.25\linewidth}
\centering
\includegraphics[width=1.45in]{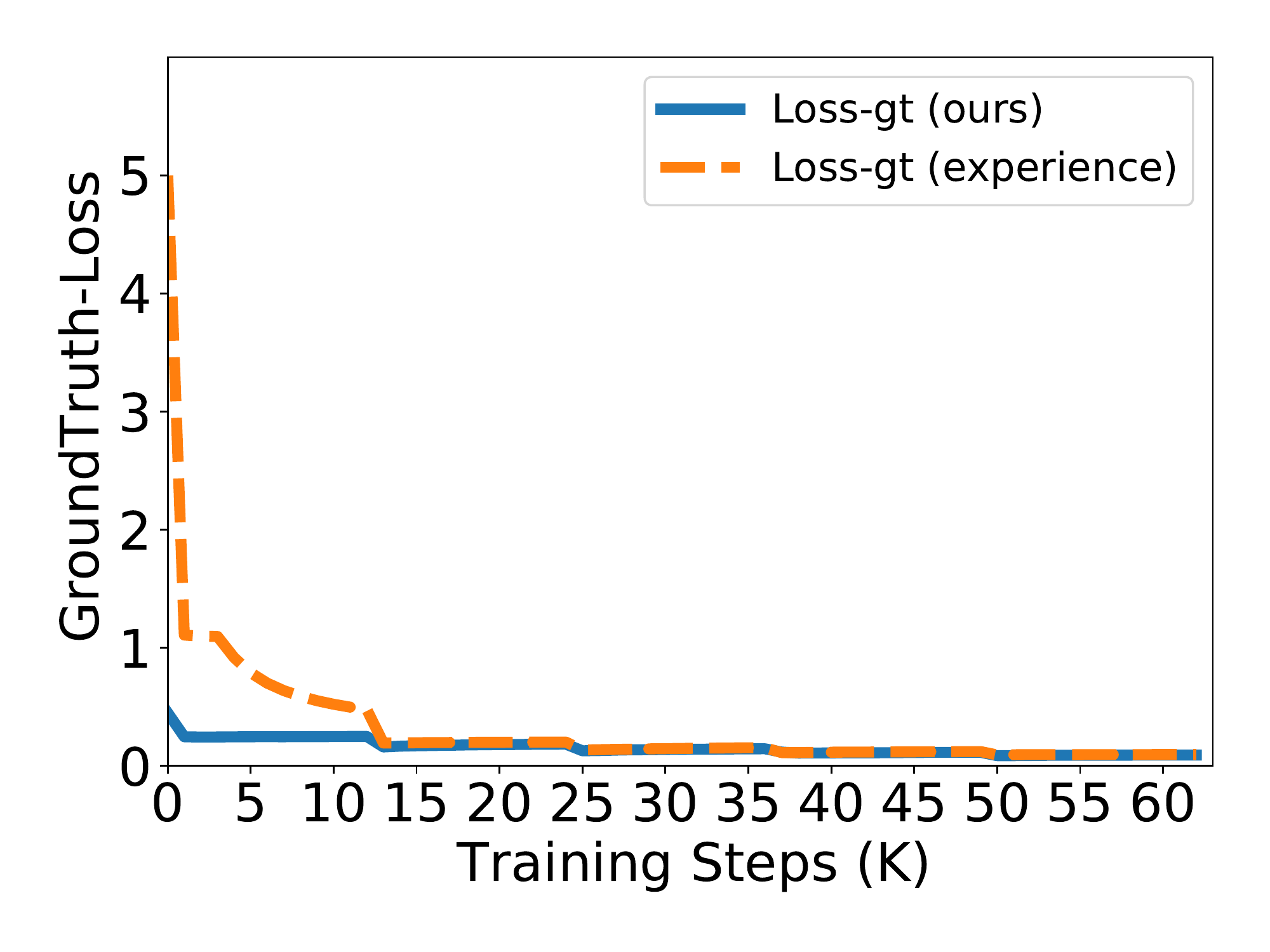}
\end{minipage}
}%
\subfigure[$Accuracy$: 2-bit]{
\begin{minipage}[t]{0.25\linewidth}
\centering
\includegraphics[width=1.45in]{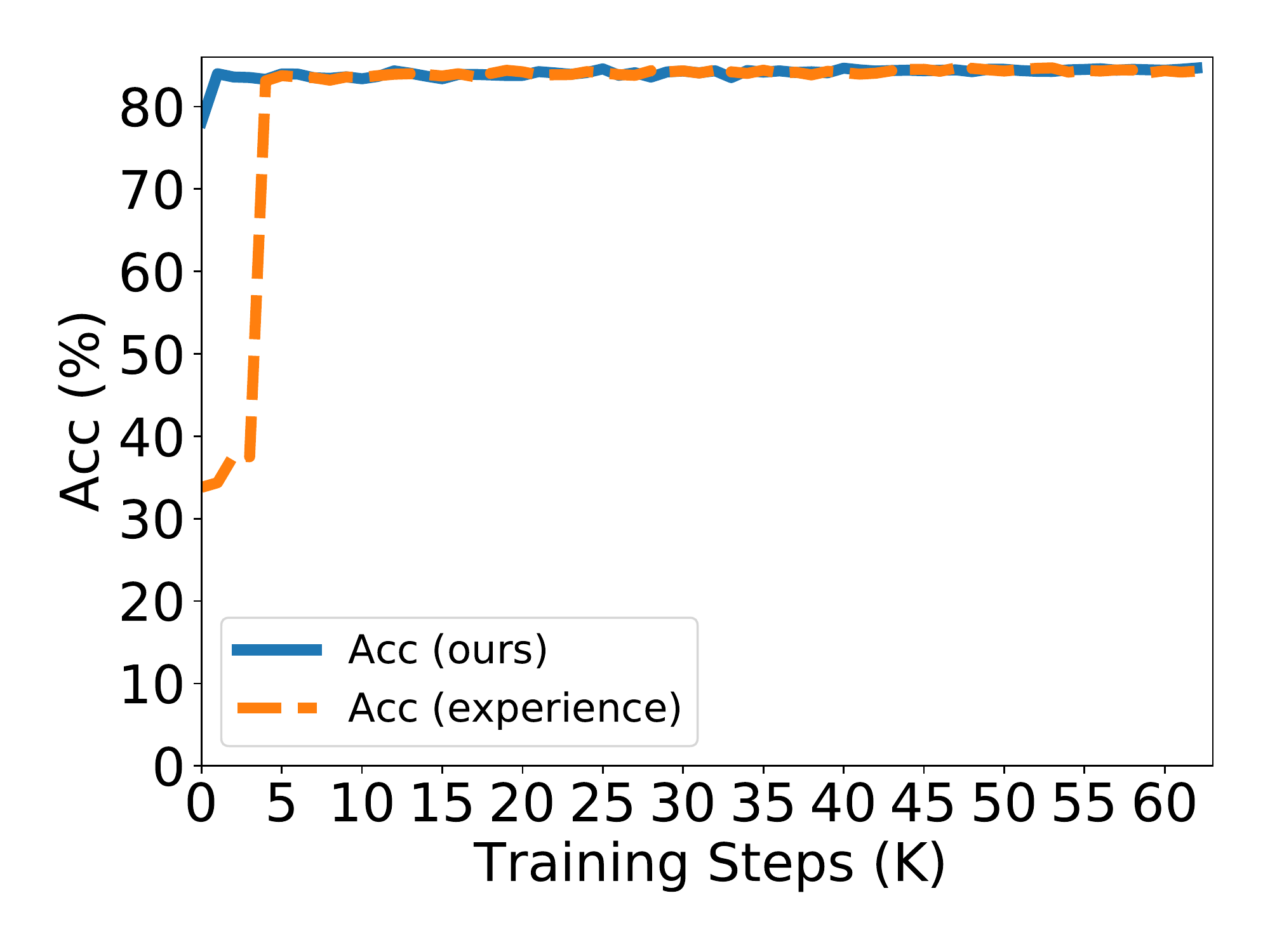}
\end{minipage}
}%
\end{center}
\caption{The impact on \textbf{total training loss},  \textbf{distillation loss}, \textbf{ground truth loss} and \textbf{accuracy}  when implementing our proposed KDLSQ-BERT by using different scale-factor initialization. \textbf{The quantization bit "W-E-A (\#bit)" is set to 2-2-8}. The experimental results are tested by adopting \textbf{"mnli"}  and \textbf{"TinyBERT"}.}
\label{figure:3}
\end{figure}

\begin{figure}[htb]
\begin{center}
\subfigure[$Loss_{total}$: 4-bit]{
\begin{minipage}[t]{0.25\linewidth}
\centering
\includegraphics[width=1.45in]{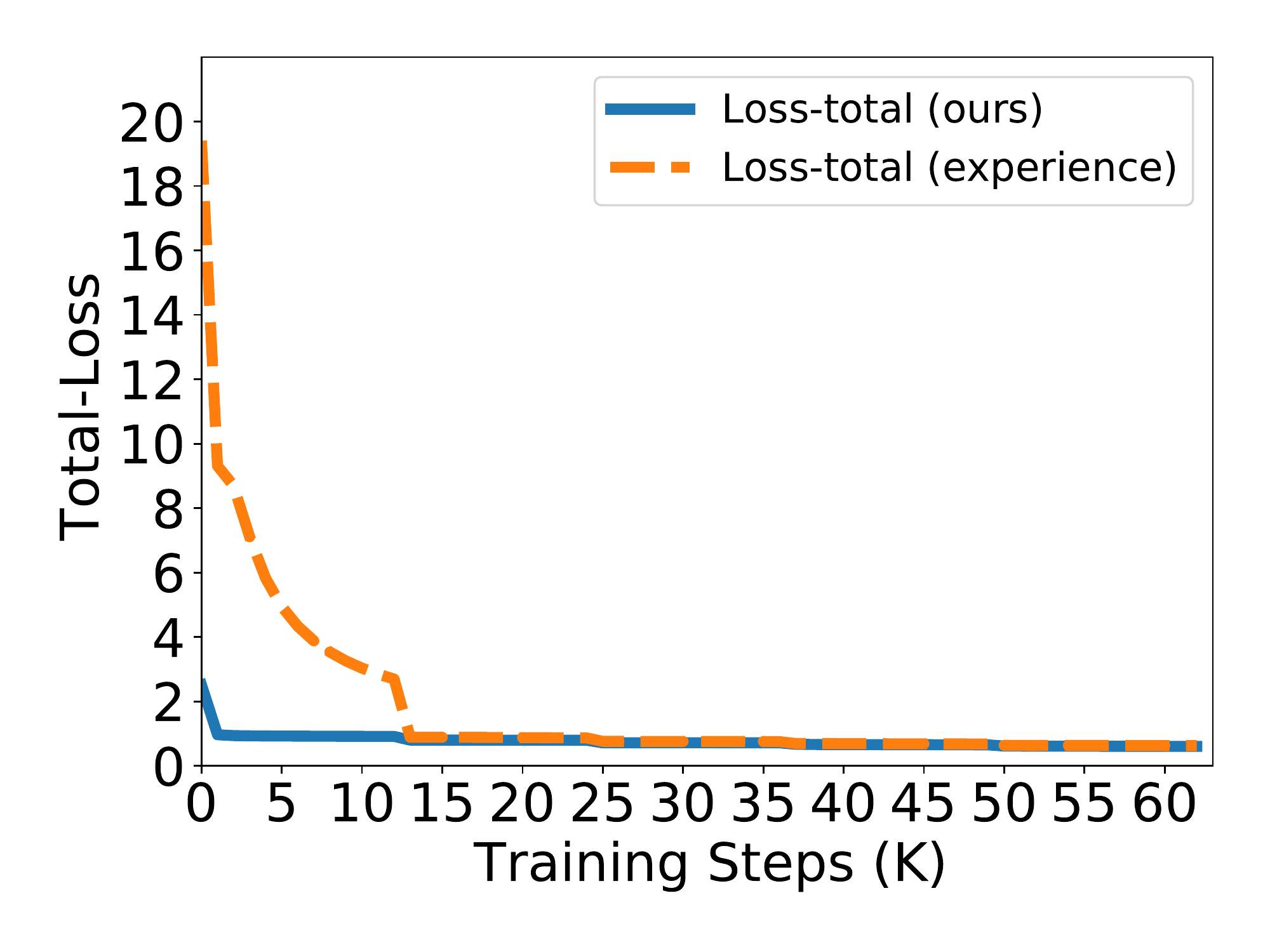}
\end{minipage}%
}%
\subfigure[$Loss_{kd}$: 4-bit]{
\begin{minipage}[t]{0.25\linewidth}
\centering
\includegraphics[width=1.45in]{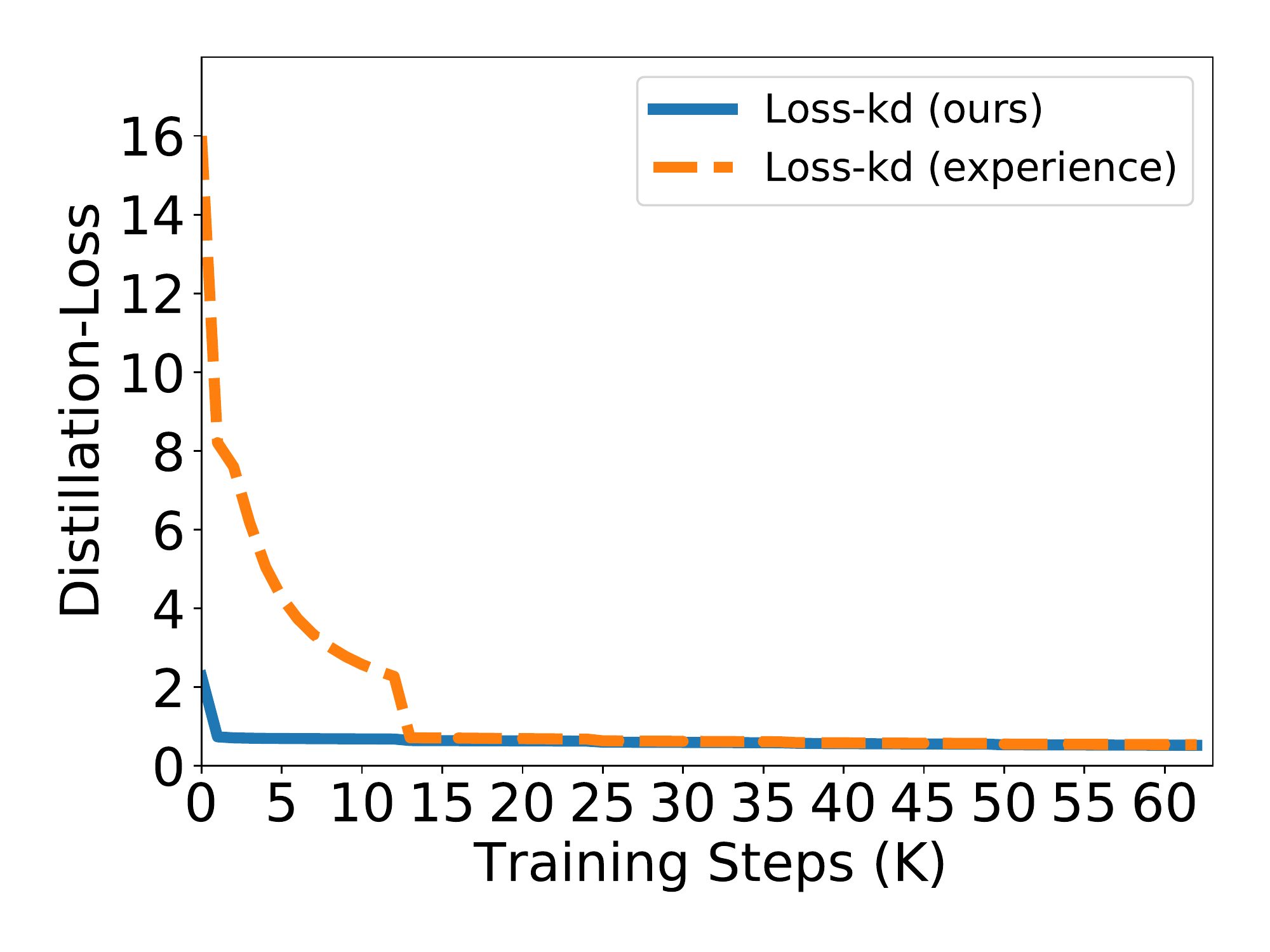}
\end{minipage}%
}%
\subfigure[$Loss_{gt}$: 4-bit]{
\begin{minipage}[t]{0.25\linewidth}
\centering
\includegraphics[width=1.45in]{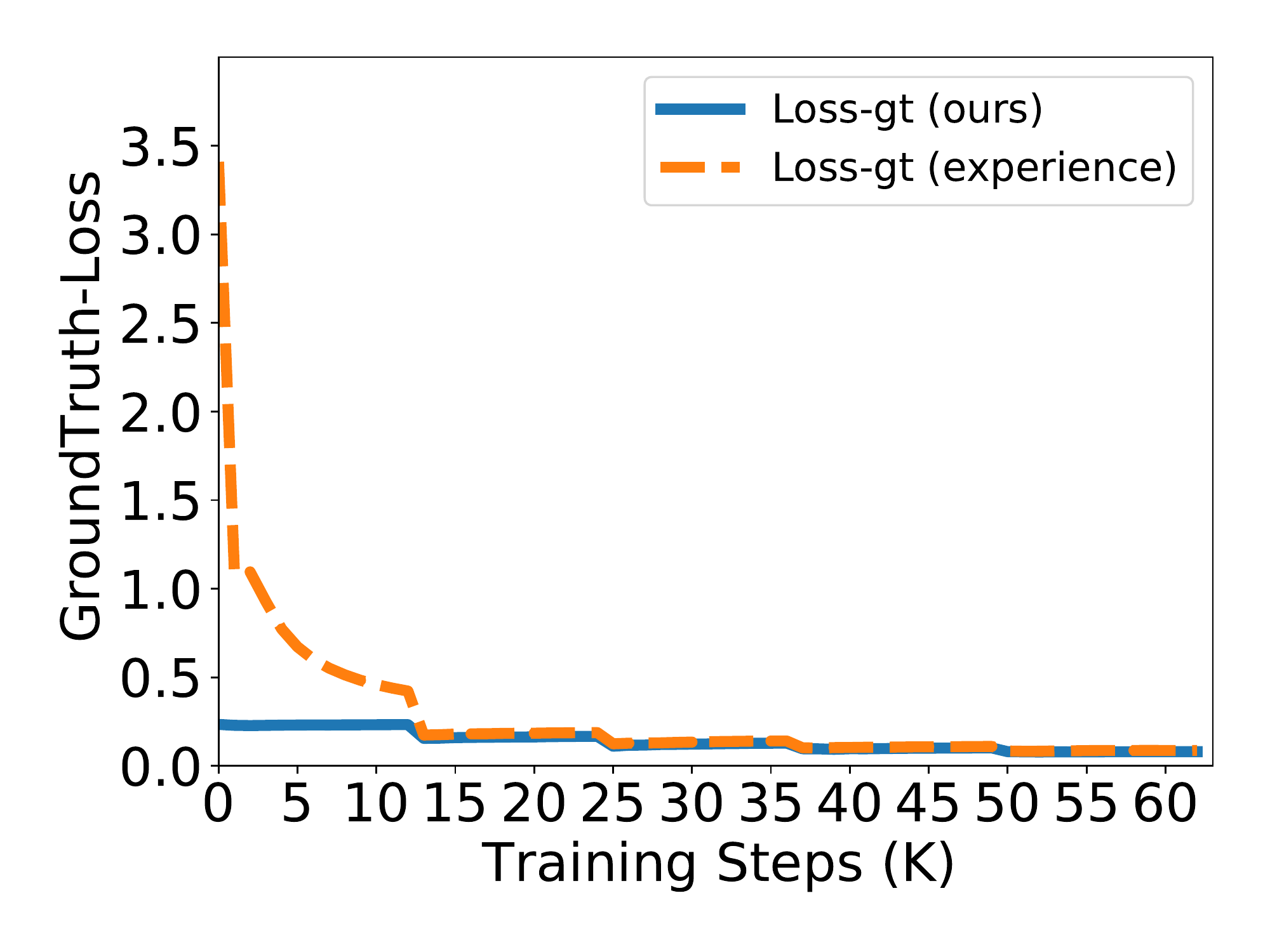}
\end{minipage}
}%
\subfigure[$Accuracy$: 4-bit]{
\begin{minipage}[t]{0.25\linewidth}
\centering
\includegraphics[width=1.45in]{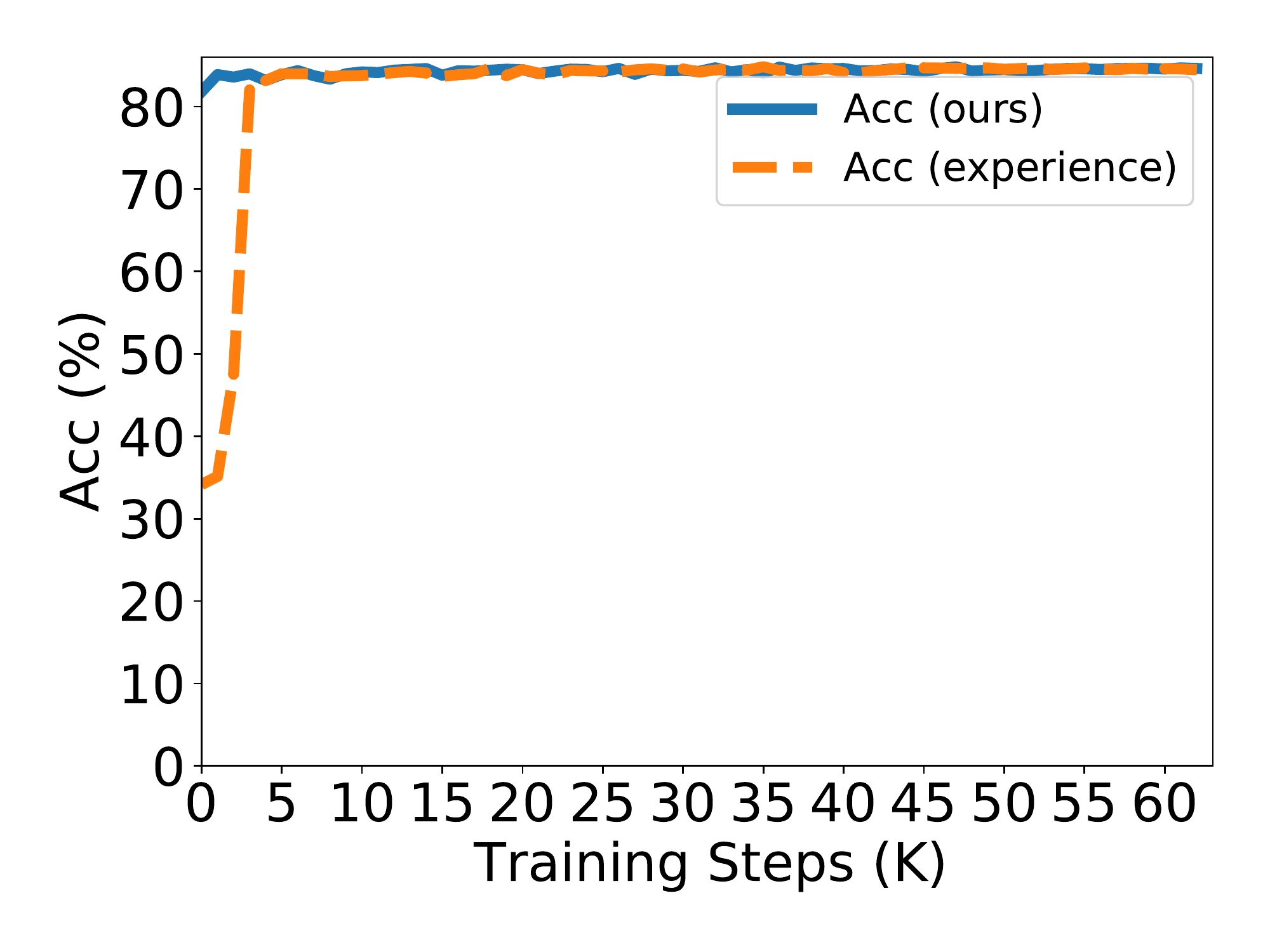}
\end{minipage}
}%
\end{center}
\caption{The impact on \textbf{total training loss},  \textbf{distillation loss}, \textbf{ground truth loss} and \textbf{accuracy}  when implementing our proposed KDLSQ-BERT by using different scale-factor initialization. \textbf{The quantization bit "W-E-A (\#bit)" is set to 4-4-8}. The experimental results are tested by adopting \textbf{"mnli"}  and \textbf{"TinyBERT"}.}
\label{figure:4}
\end{figure}

\begin{figure}[htb]
\begin{center}
\subfigure[$Loss_{total}$: 6-bit]{
\begin{minipage}[t]{0.25\linewidth}
\centering
\includegraphics[width=1.45in]{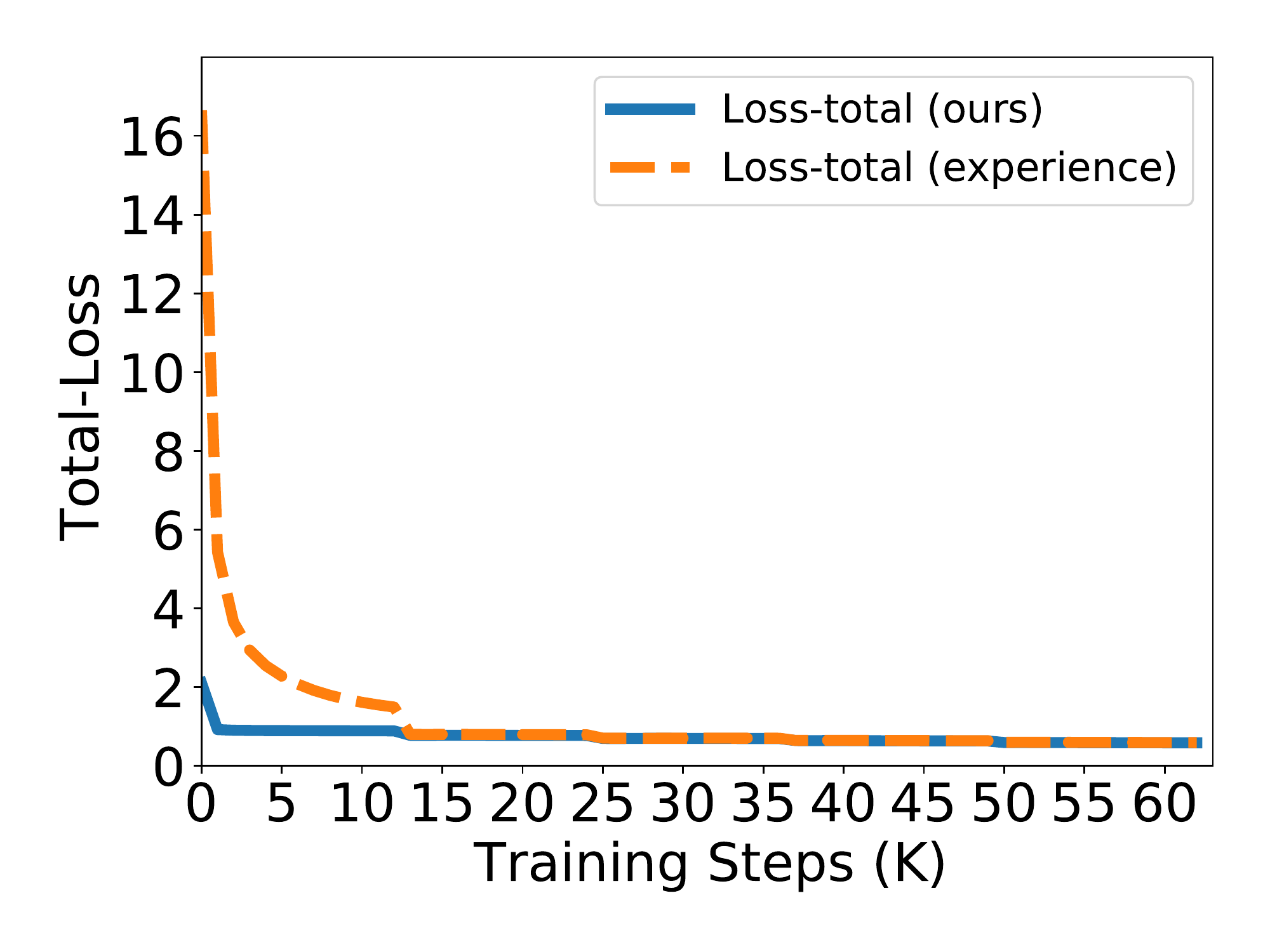}
\end{minipage}%
}%
\subfigure[$Loss_{kd}$: 6-bit]{
\begin{minipage}[t]{0.25\linewidth}
\centering
\includegraphics[width=1.45in]{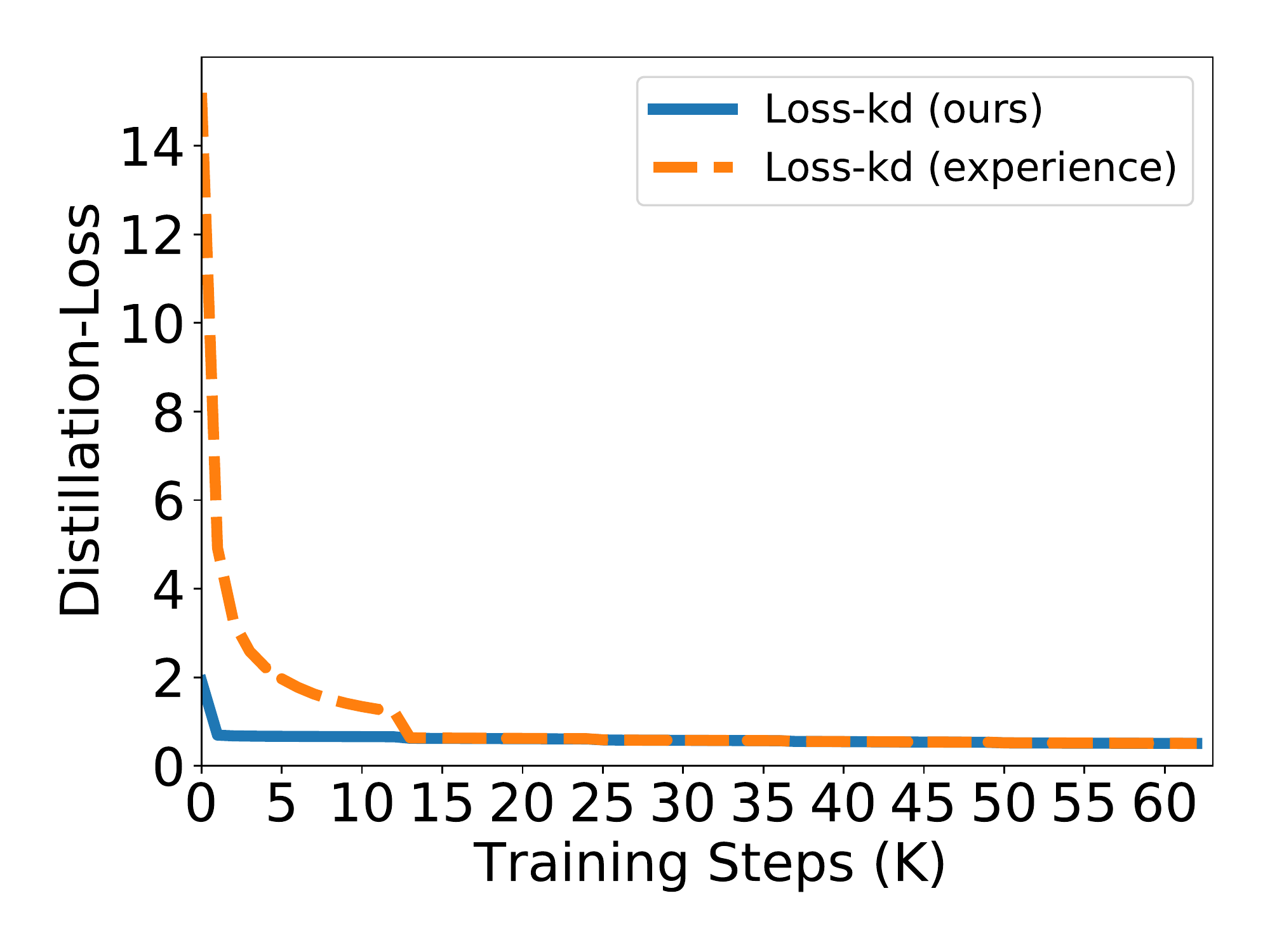}
\end{minipage}%
}%
\subfigure[$Loss_{gt}$: 6-bit]{
\begin{minipage}[t]{0.25\linewidth}
\centering
\includegraphics[width=1.45in]{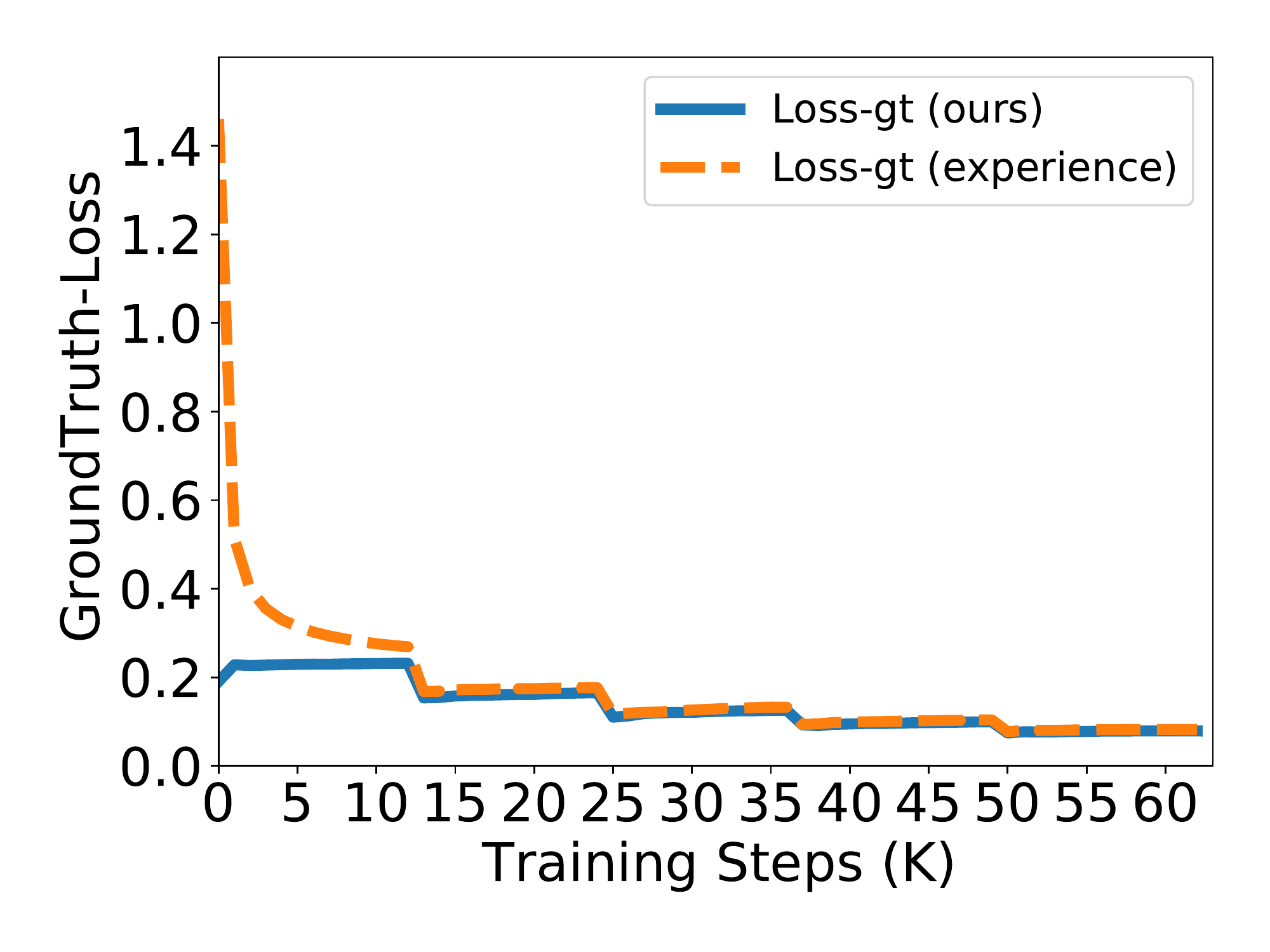}
\end{minipage}
}%
\subfigure[$Accuracy$: 6-bit]{
\begin{minipage}[t]{0.25\linewidth}
\centering
\includegraphics[width=1.45in]{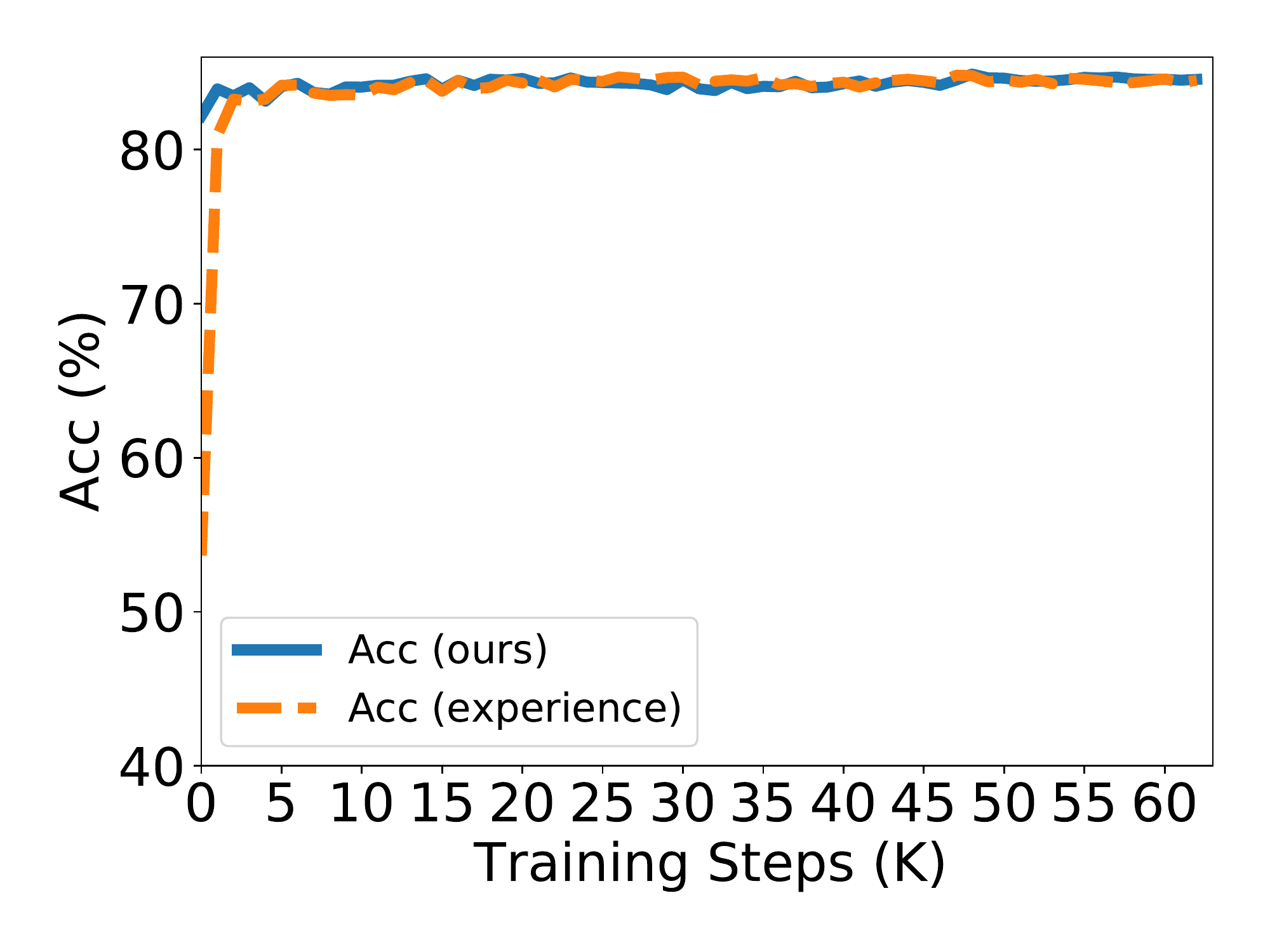}
\end{minipage}
}%
\end{center}
\caption{The impact on \textbf{total training loss},  \textbf{distillation loss}, \textbf{ground truth loss} and \textbf{accuracy}  when implementing our proposed KDLSQ-BERT by using different scale-factor initialization.\textbf{The quantization bit "W-E-A (\#bit)" is set to 6-6-8}. The experimental results are tested by adopting \textbf{"mnli"}  and \textbf{"TinyBERT"}.}
\label{figure:5}
\end{figure}

\begin{figure}[htb]
\begin{center}
\subfigure[$Loss_{total}$: 8-bit]{
\begin{minipage}[t]{0.25\linewidth}
\centering
\includegraphics[width=1.45in]{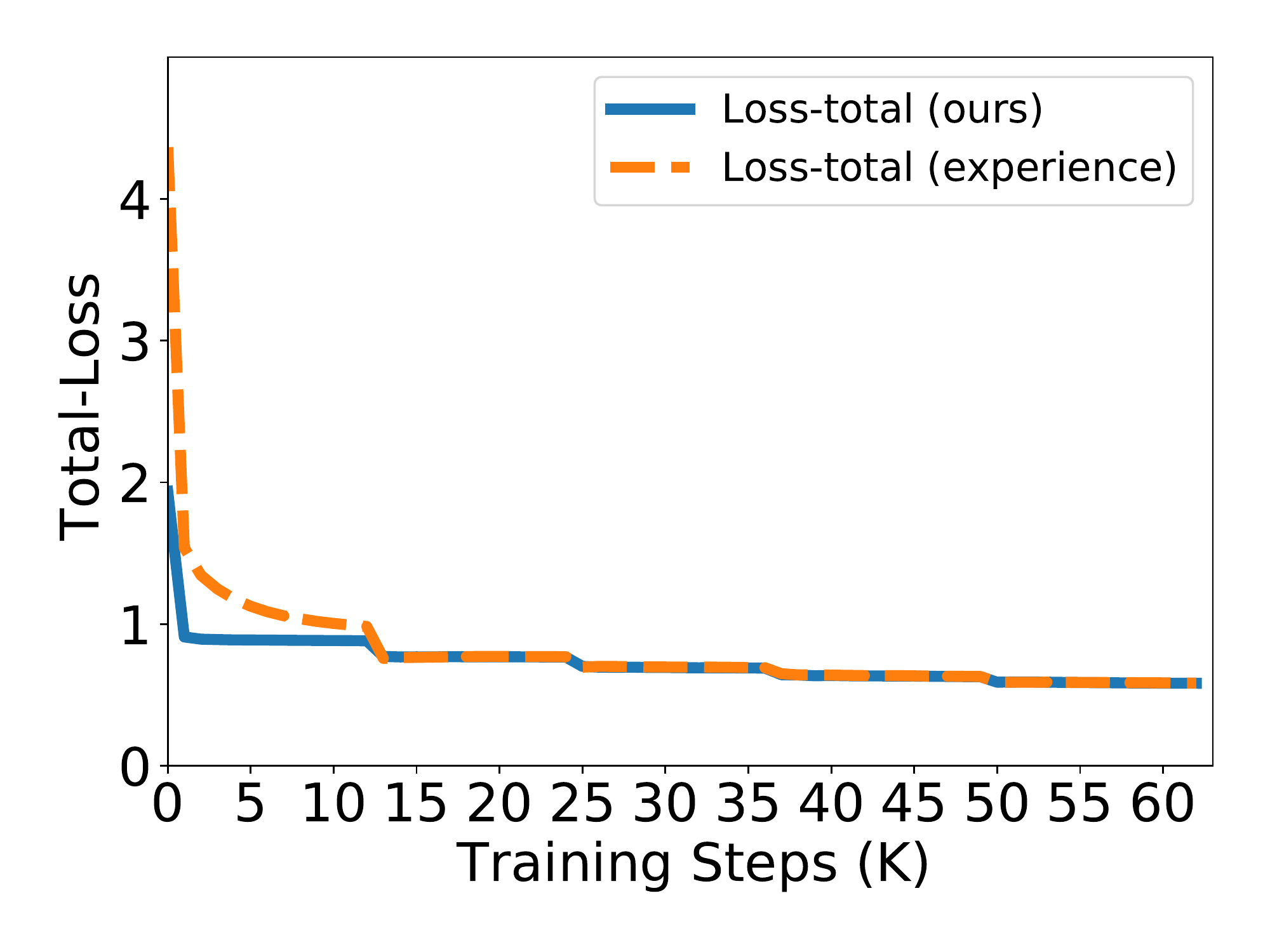}
\end{minipage}%
}%
\subfigure[$Loss_{kd}$: 8-bit]{
\begin{minipage}[t]{0.25\linewidth}
\centering
\includegraphics[width=1.45in]{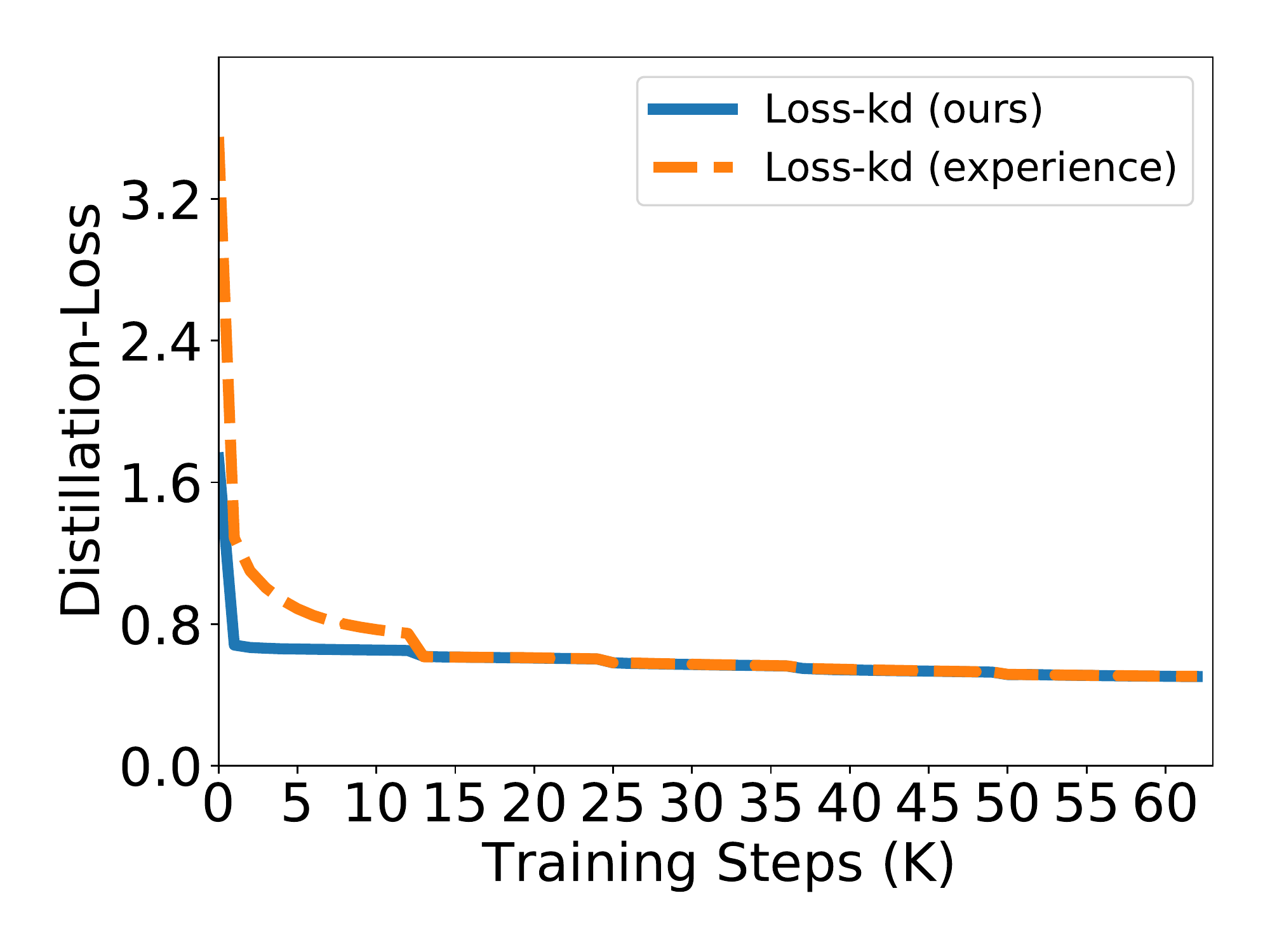}
\end{minipage}%
}%
\subfigure[$Loss_{gt}$: 8-bit]{
\begin{minipage}[t]{0.25\linewidth}
\centering
\includegraphics[width=1.45in]{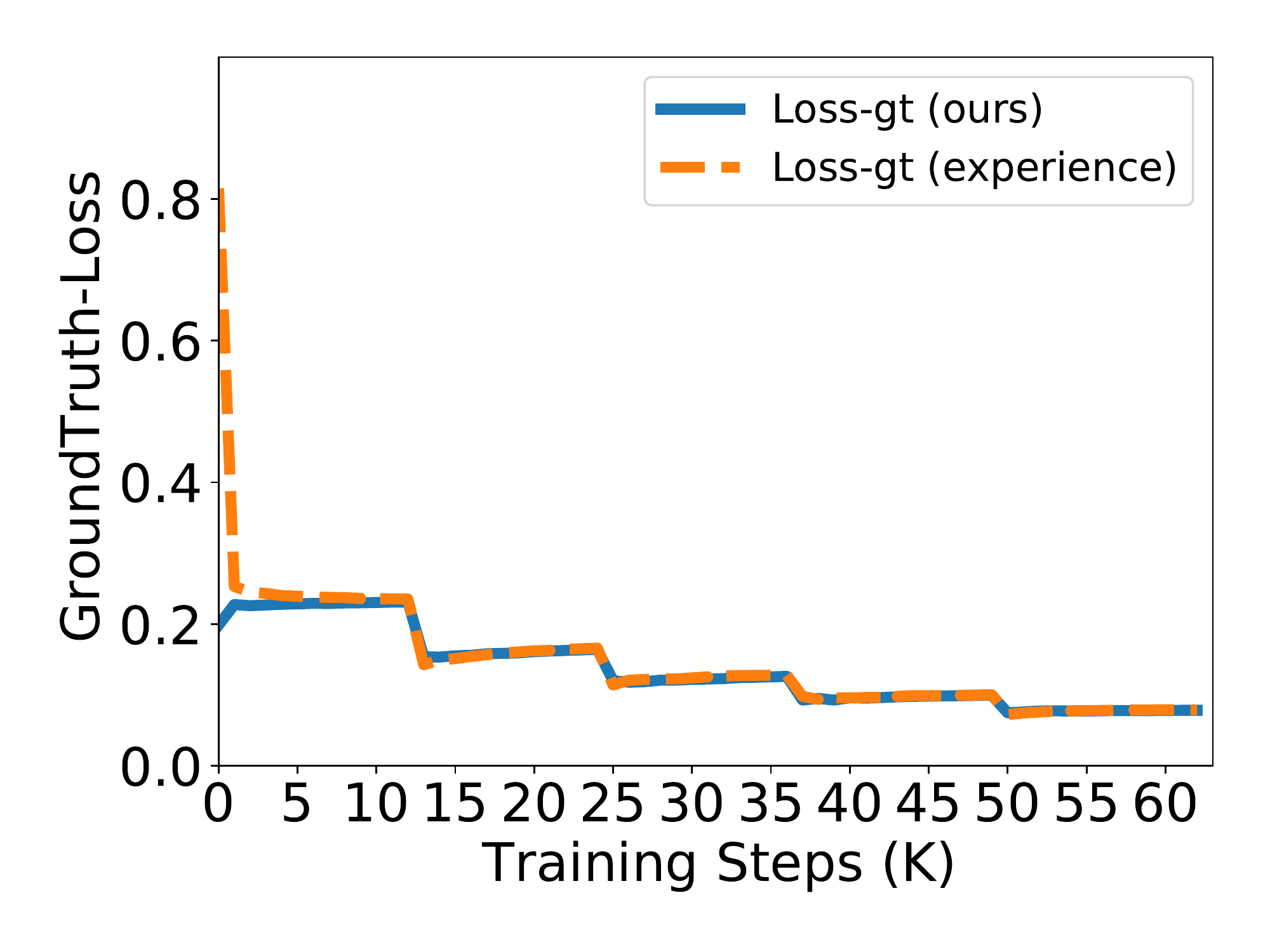}
\end{minipage}
}%
\subfigure[$Accuracy$: 8-bit]{
\begin{minipage}[t]{0.25\linewidth}
\centering
\includegraphics[width=1.45in]{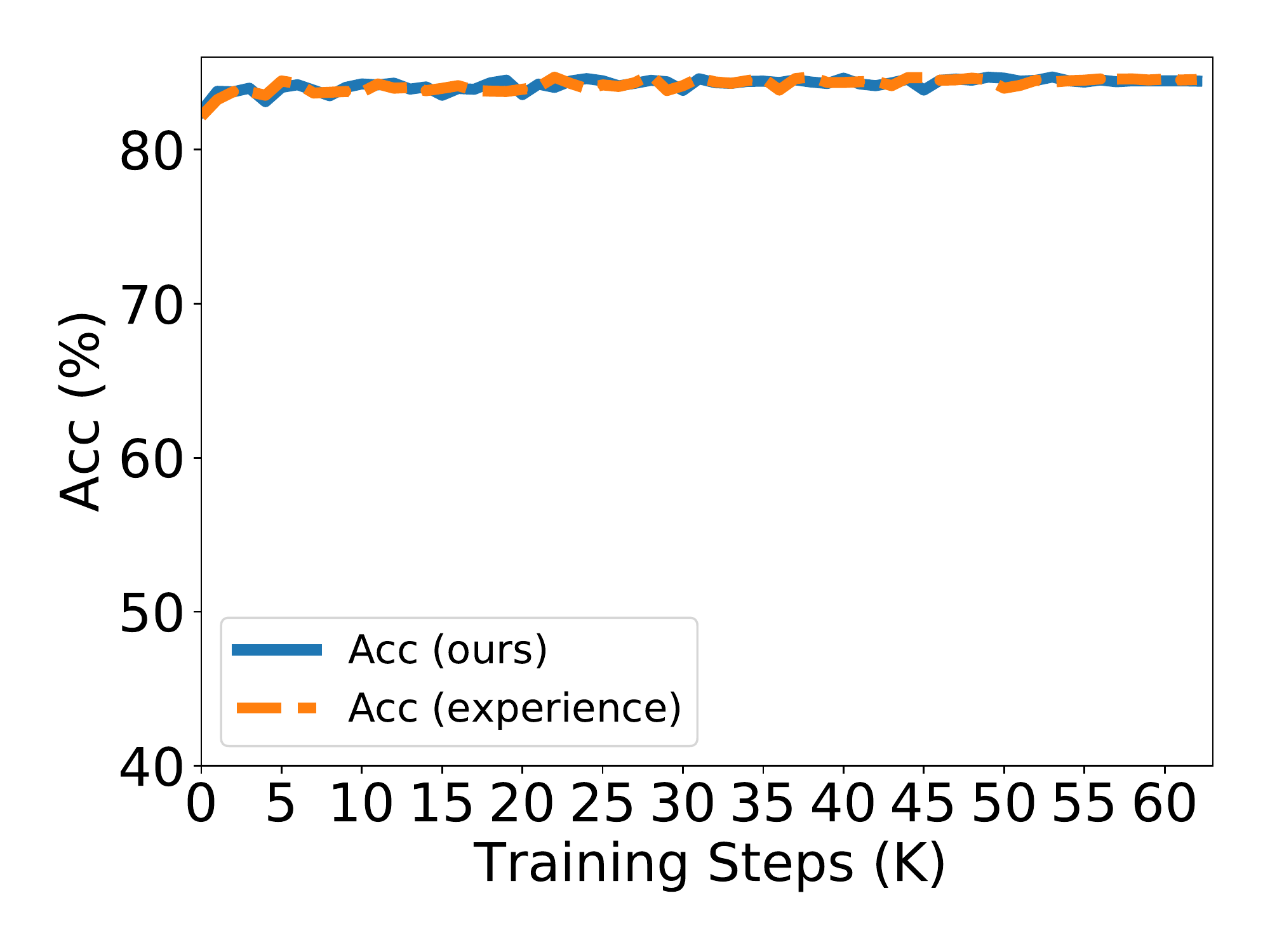}
\end{minipage}
}%
\end{center}
\caption{The impact on \textbf{total training loss},  \textbf{distillation loss}, \textbf{ground truth loss} and \textbf{accuracy}  when implementing our proposed KDLSQ-BERT by using different scale-factor initialization. \textbf{The quantization bit "W-E-A (\#bit)" is set to 8-8-8}. The experimental results are tested by adopting \textbf{"mnli"}  and \textbf{"TinyBERT"}.}
\label{figure:6}
\end{figure}

From Figure \ref{figure:3} $\sim$ Figure \ref{figure:6}, it illustrates the experimental results related to "mnli". As can be seen that compared with the experience value for scale-factor initialization, the correlated training loss and model accuracy have faster convergence speed when applying Algorithm \ref{alg:scale_factor_initialization} to do scale-factor initialization. The main reason is that Algorithm \ref{alg:scale_factor_initialization} can provide well initialized scale-factors for quantization training, such that a tensor that needs to be quantized can be retained main information after the truncation. Also, note that the results from the figures show other two important information. One is that all initialized training losses (that's, the initialized $Loss_{total}$, $Loss_{kd}$ and $Loss_{gt}$) decrease as the weight quantization bit increases. It implies that bigger quantization bit is good to accelerate the training convergence. The other one is that all correlated training losses can be reduced effectively through the training process, which further reveals that combining $Loss_{kd}$ with $Loss_{gt}$ for training is good to improve the model accuracy performance.

\begin{figure}[htb]
\begin{center}
\subfigure[$Loss_{total}$: 2-bit]{
\begin{minipage}[t]{0.25\linewidth}
\centering
\includegraphics[width=1.45in]{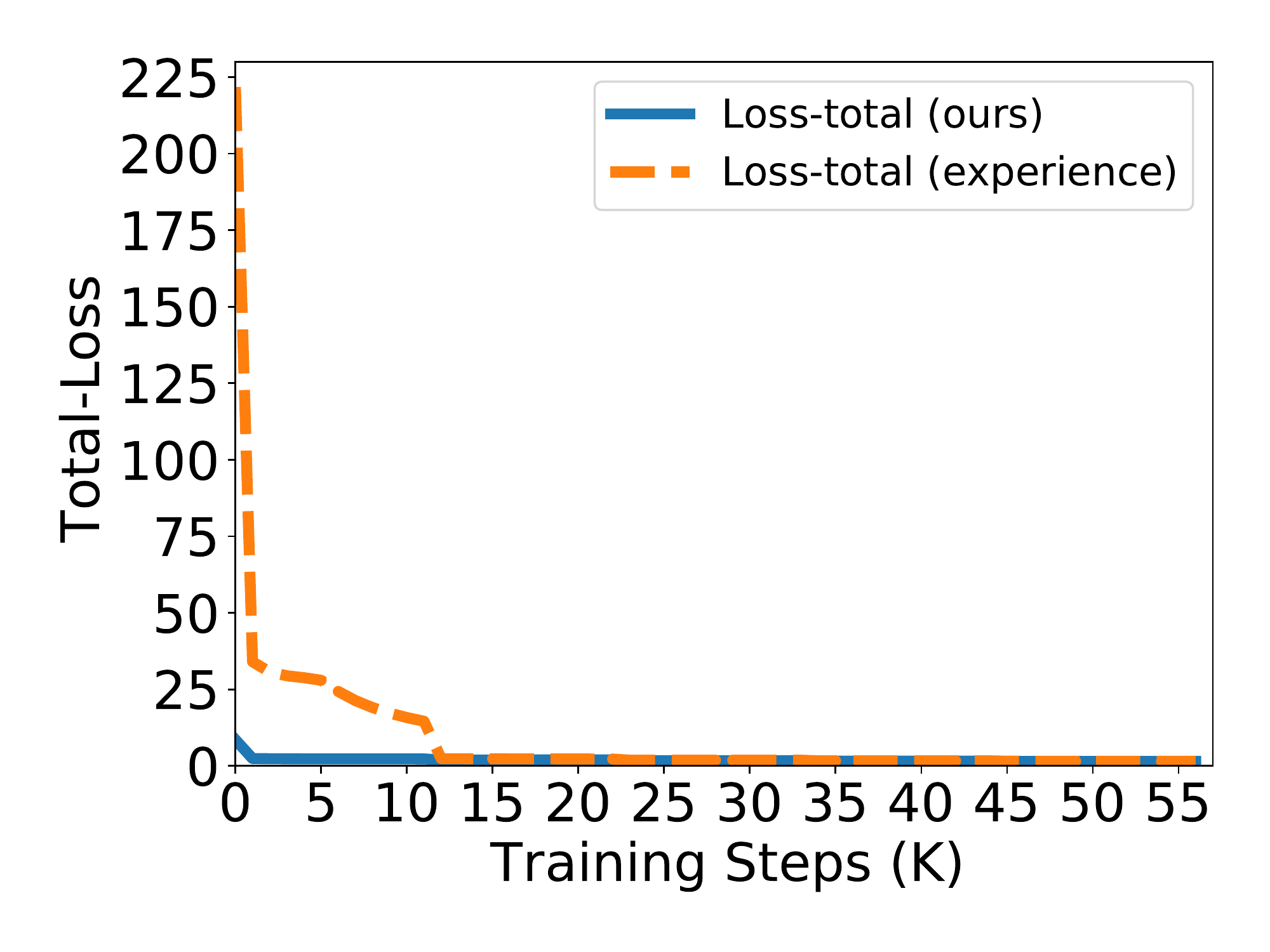}
\end{minipage}%
}%
\subfigure[$Loss_{kd}$: 2-bit]{
\begin{minipage}[t]{0.25\linewidth}
\centering
\includegraphics[width=1.45in]{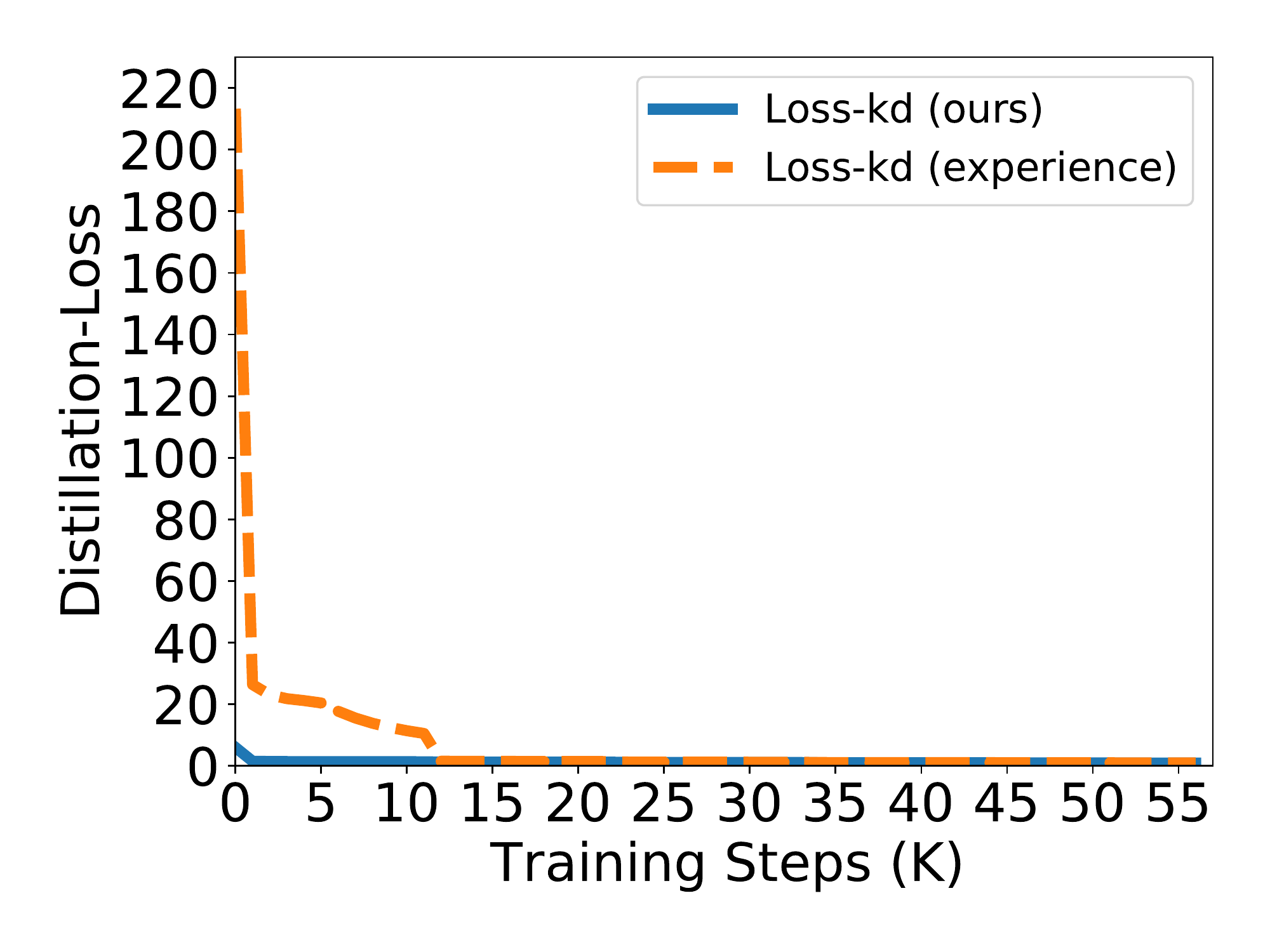}
\end{minipage}%
}%
\subfigure[$Loss_{gt}$: 2-bit]{
\begin{minipage}[t]{0.25\linewidth}
\centering
\includegraphics[width=1.45in]{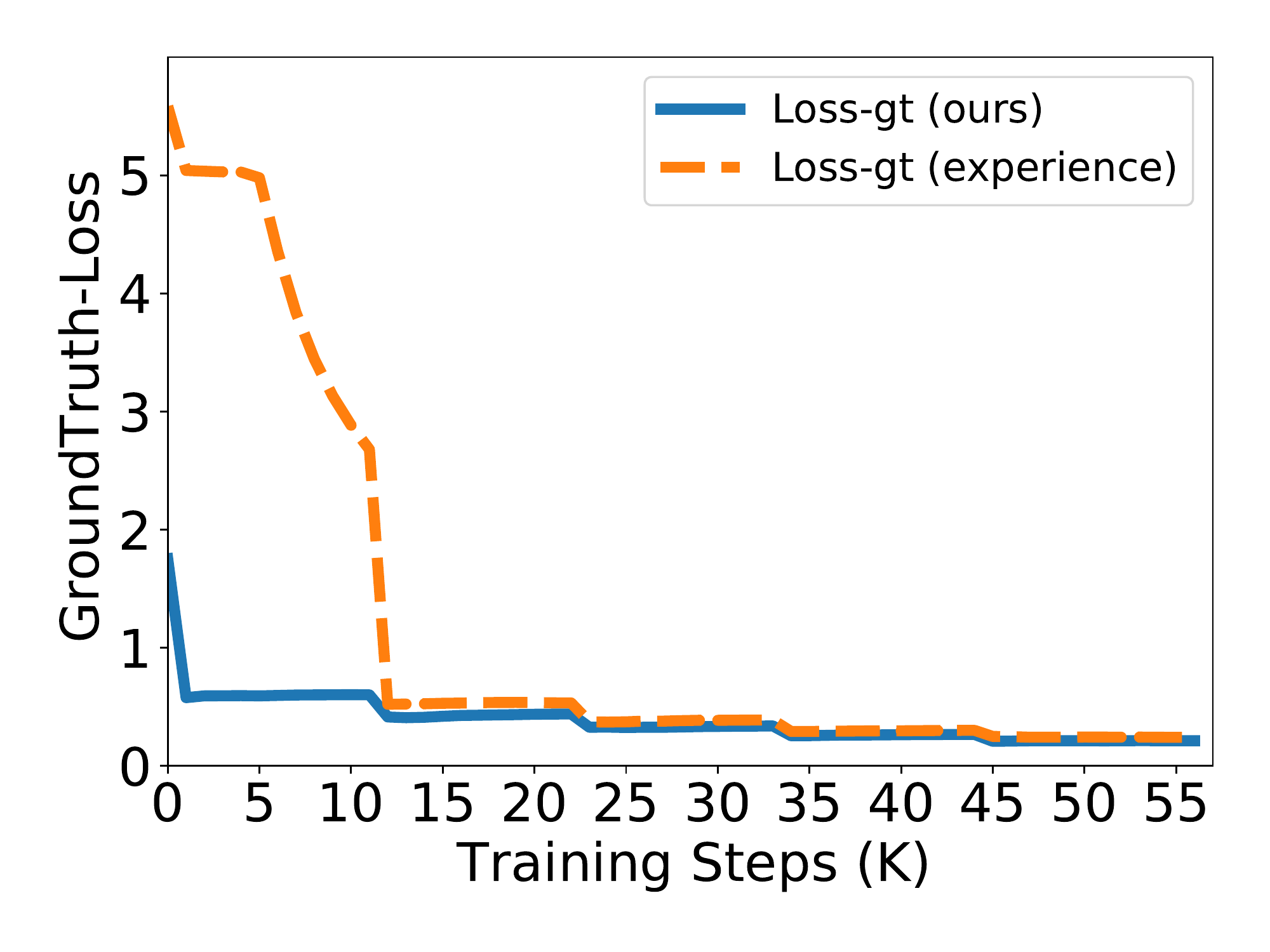}
\end{minipage}
}%
\subfigure[$F1$: 2-bit]{
\begin{minipage}[t]{0.25\linewidth}
\centering
\includegraphics[width=1.45in]{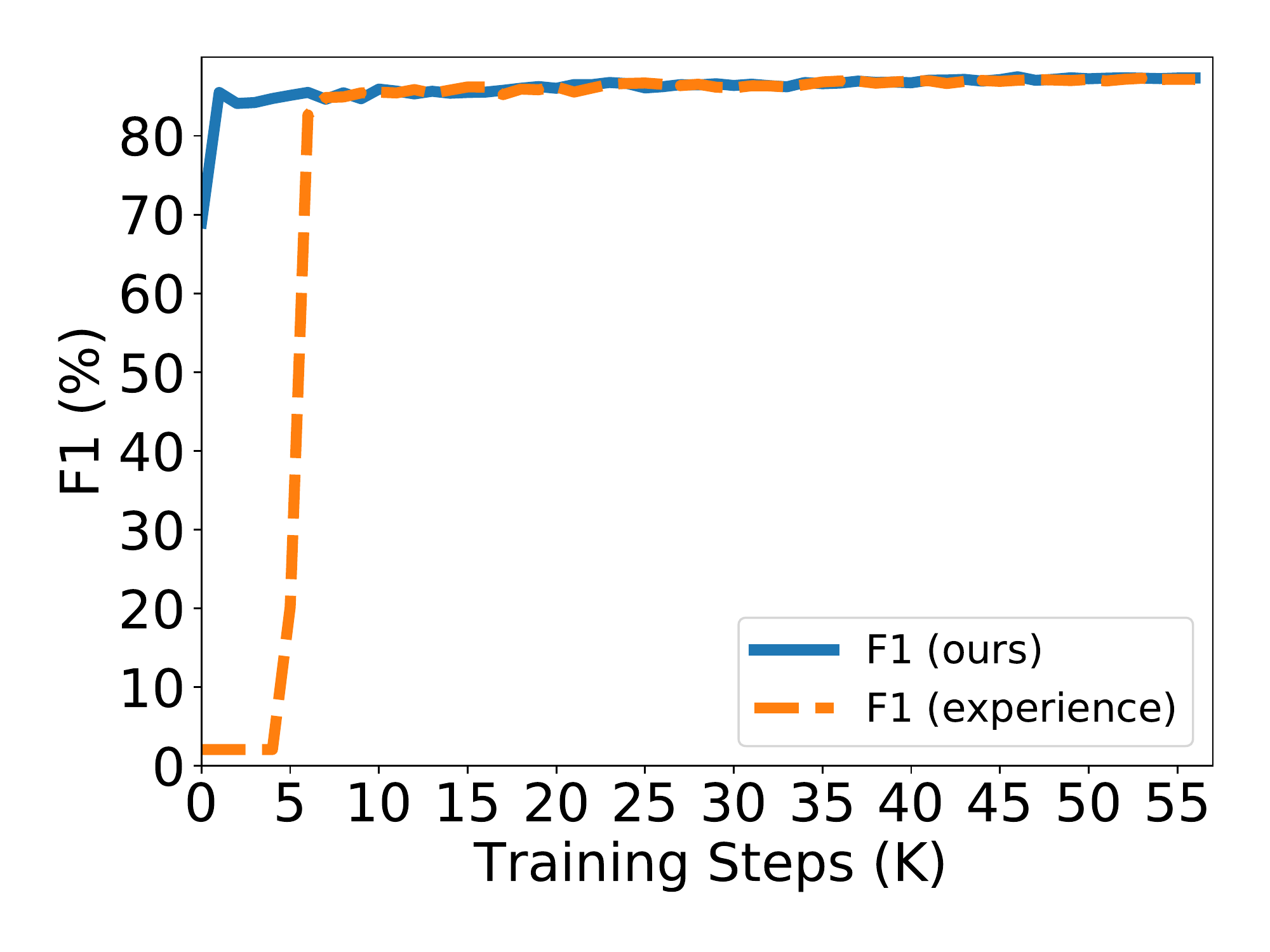}
\end{minipage}
}%
\end{center}
\caption{The impact on \textbf{total training loss},  \textbf{distillation loss}, \textbf{ground truth loss} and \textbf{accuracy}  when implementing our proposed KDLSQ-BERT by using different scale-factor initialization. \textbf{The quantization bit "W-E-A (\#bit)" is set to 2-2-8}. The experimental results are tested by adopting \textbf{"SQuAD 1.1"}  and \textbf{"TinyBERT"}.}
\label{figure:7}
\end{figure}

\begin{figure}[htb]
\begin{center}
\subfigure[$Loss_{total}$: 4-bit]{
\begin{minipage}[t]{0.25\linewidth}
\centering
\includegraphics[width=1.45in]{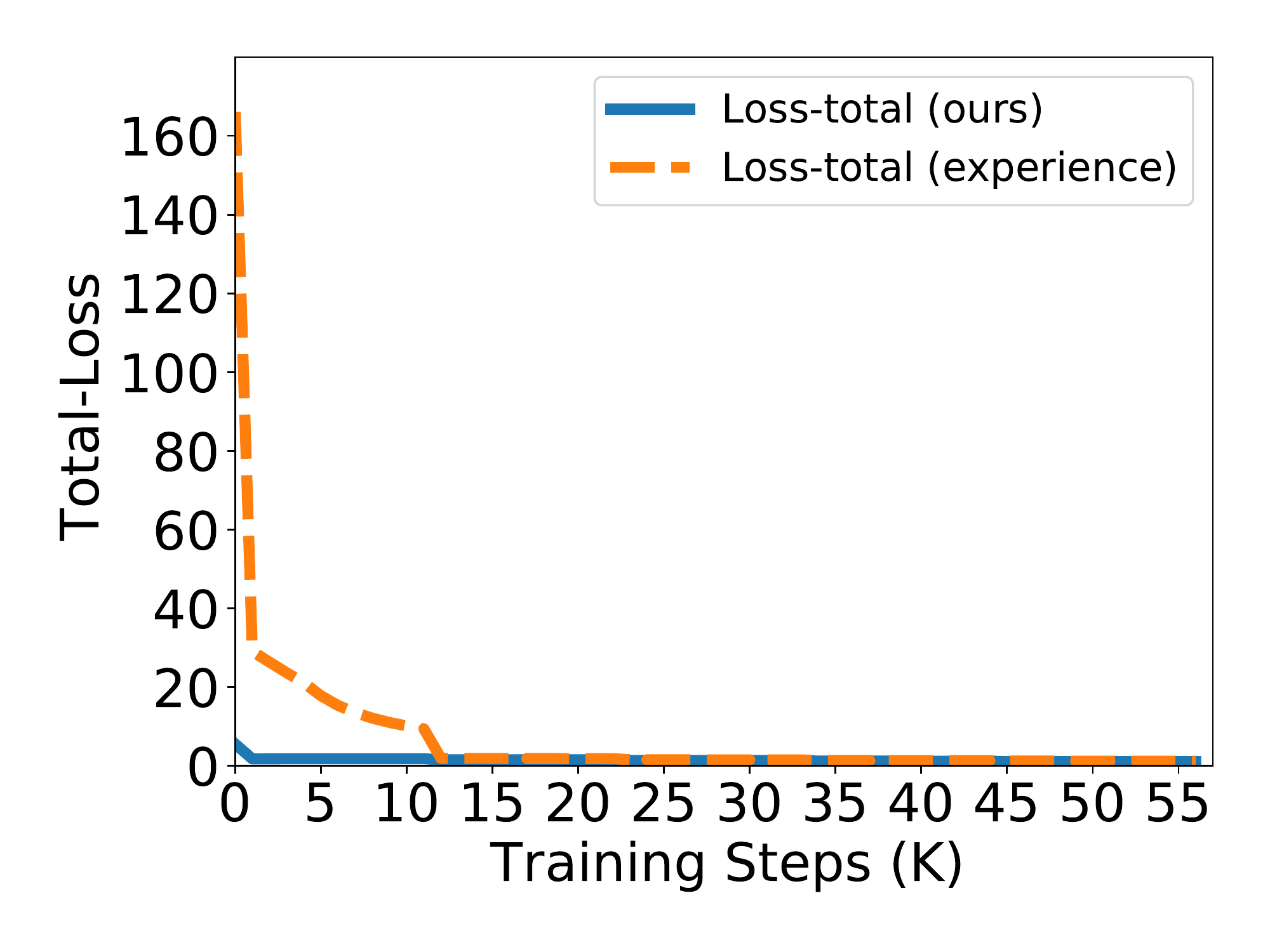}
\end{minipage}%
}%
\subfigure[$Loss_{kd}$: 4-bit]{
\begin{minipage}[t]{0.25\linewidth}
\centering
\includegraphics[width=1.45in]{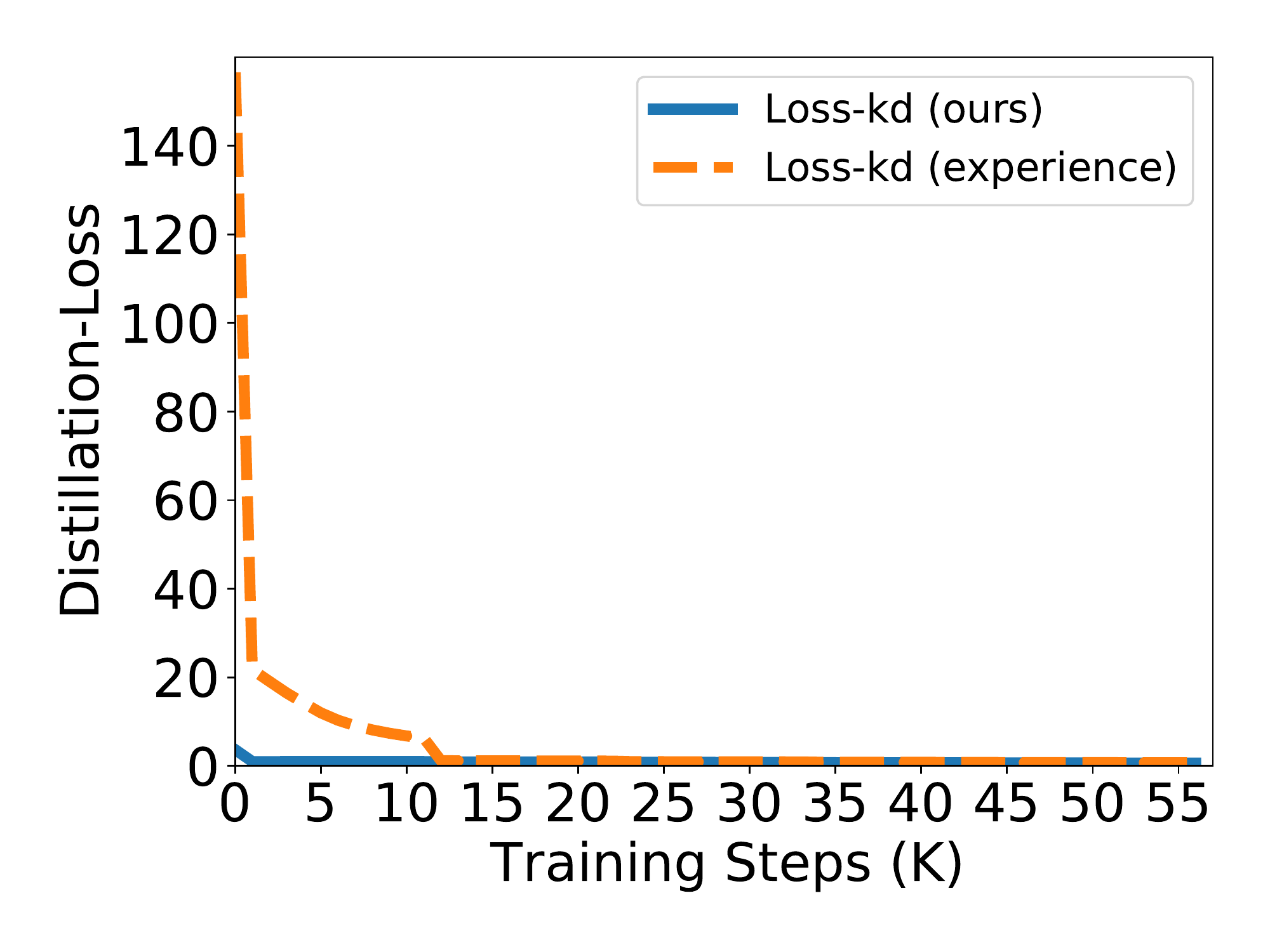}
\end{minipage}%
}%
\subfigure[$Loss_{gt}$: 4-bit]{
\begin{minipage}[t]{0.25\linewidth}
\centering
\includegraphics[width=1.45in]{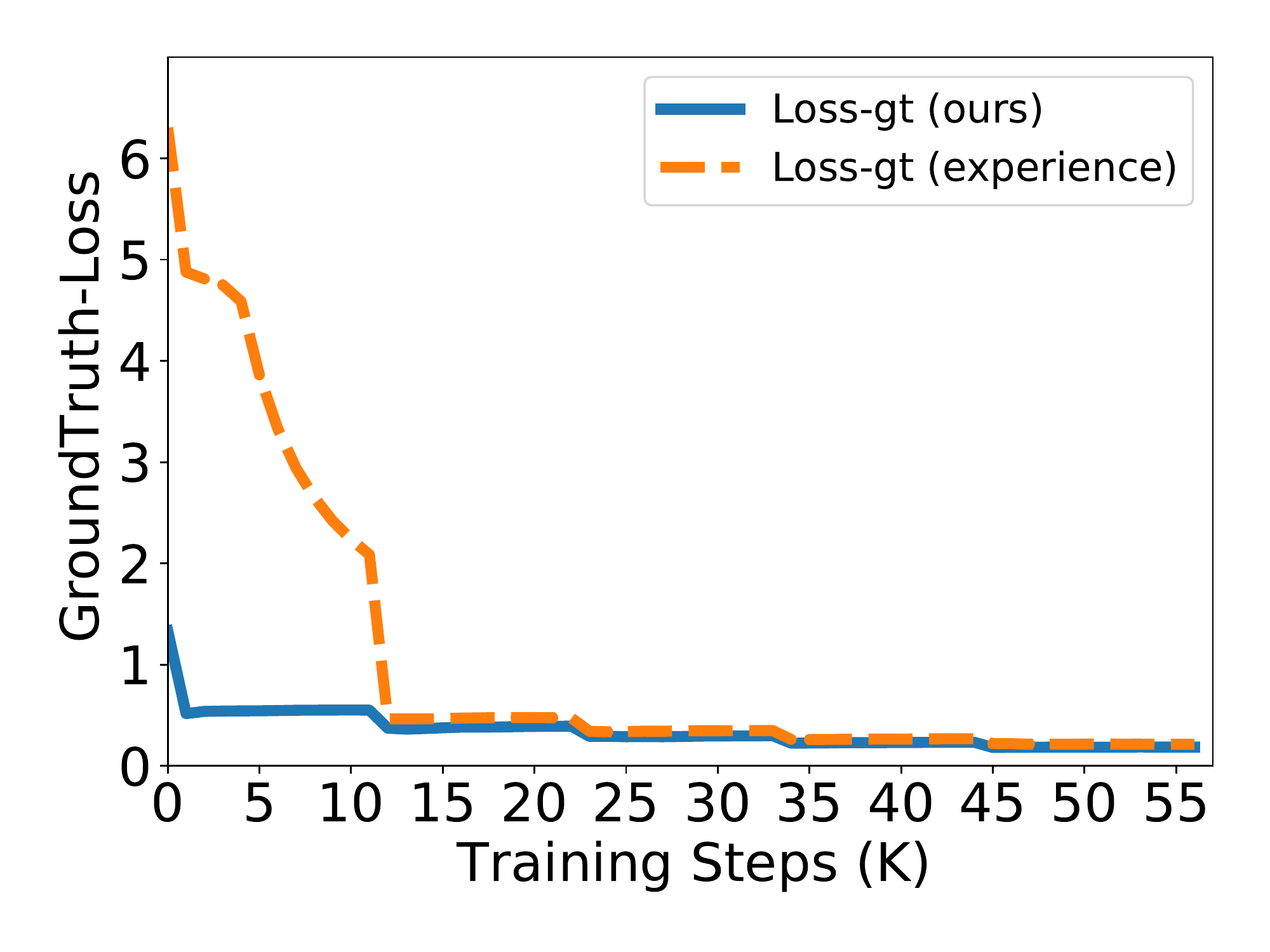}
\end{minipage}
}%
\subfigure[$F1$: 4-bit]{
\begin{minipage}[t]{0.25\linewidth}
\centering
\includegraphics[width=1.45in]{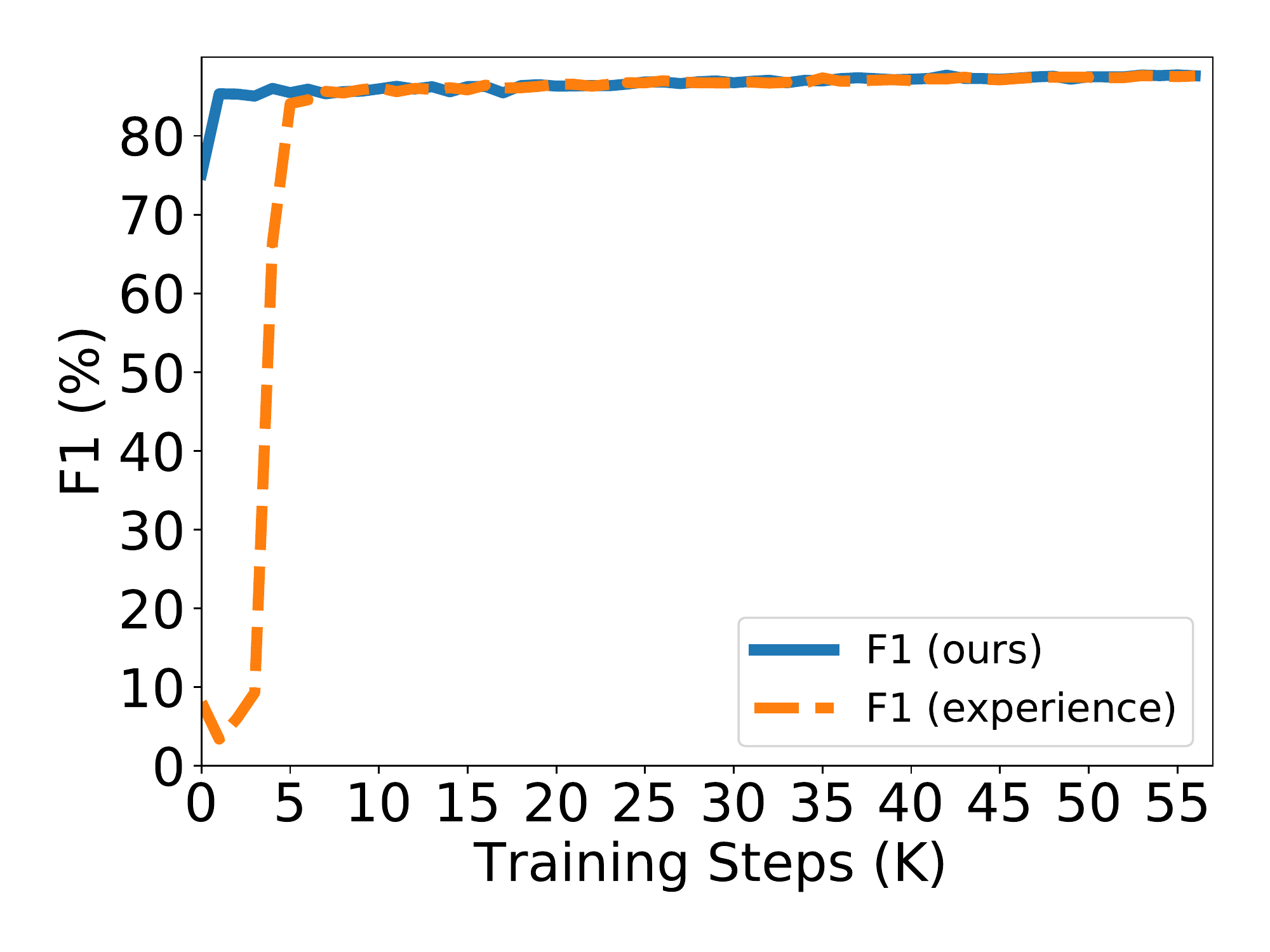}
\end{minipage}
}%
\end{center}
\caption{The impact on \textbf{total training loss},  \textbf{distillation loss}, \textbf{ground truth loss} and \textbf{accuracy}  when implementing our proposed KDLSQ-BERT by using different scale-factor initialization. \textbf{The quantization bit "W-E-A (\#bit)" is set to 4-4-8}. The experimental results are tested by adopting \textbf{"SQuAD 1.1"}  and \textbf{"TinyBERT"}}
\label{figure:8}
\end{figure}

\begin{figure}[htb]
\begin{center}
\subfigure[$Loss_{total}$: 6-bit]{
\begin{minipage}[t]{0.25\linewidth}
\centering
\includegraphics[width=1.45in]{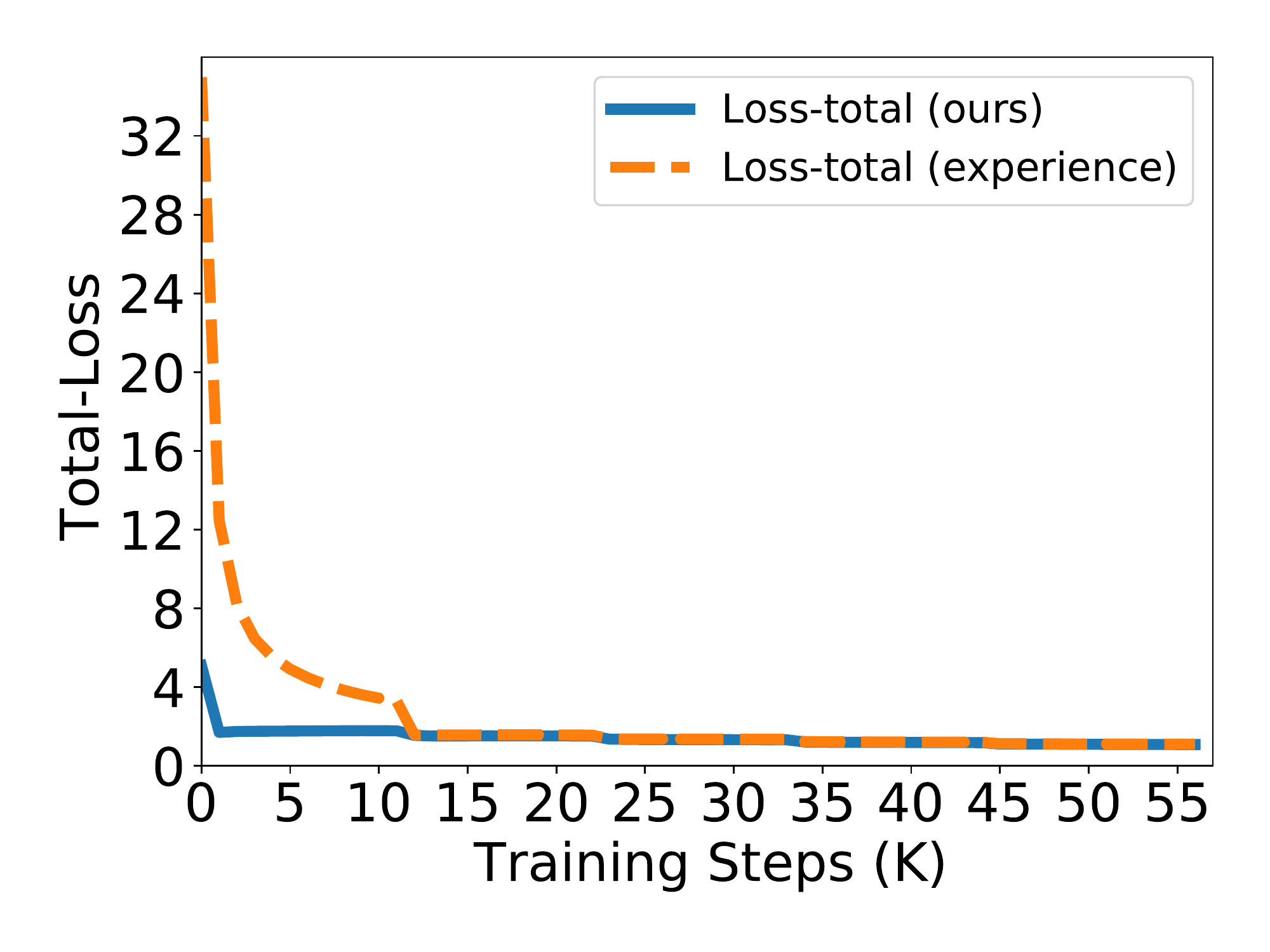}
\end{minipage}%
}%
\subfigure[$Loss_{kd}$: 6-bit]{
\begin{minipage}[t]{0.25\linewidth}
\centering
\includegraphics[width=1.45in]{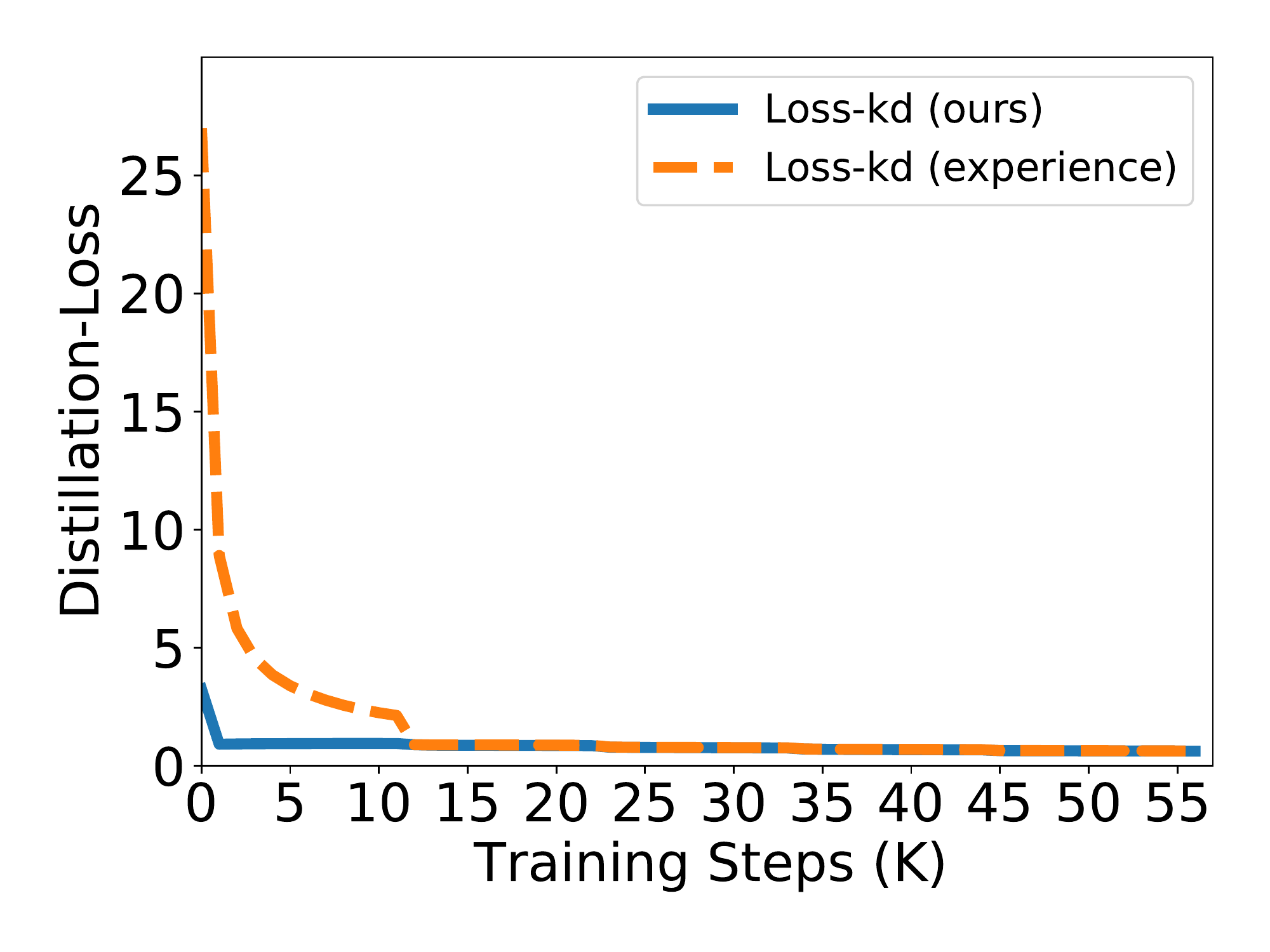}
\end{minipage}%
}%
\subfigure[$Loss_{gt}$: 6-bit]{
\begin{minipage}[t]{0.25\linewidth}
\centering
\includegraphics[width=1.45in]{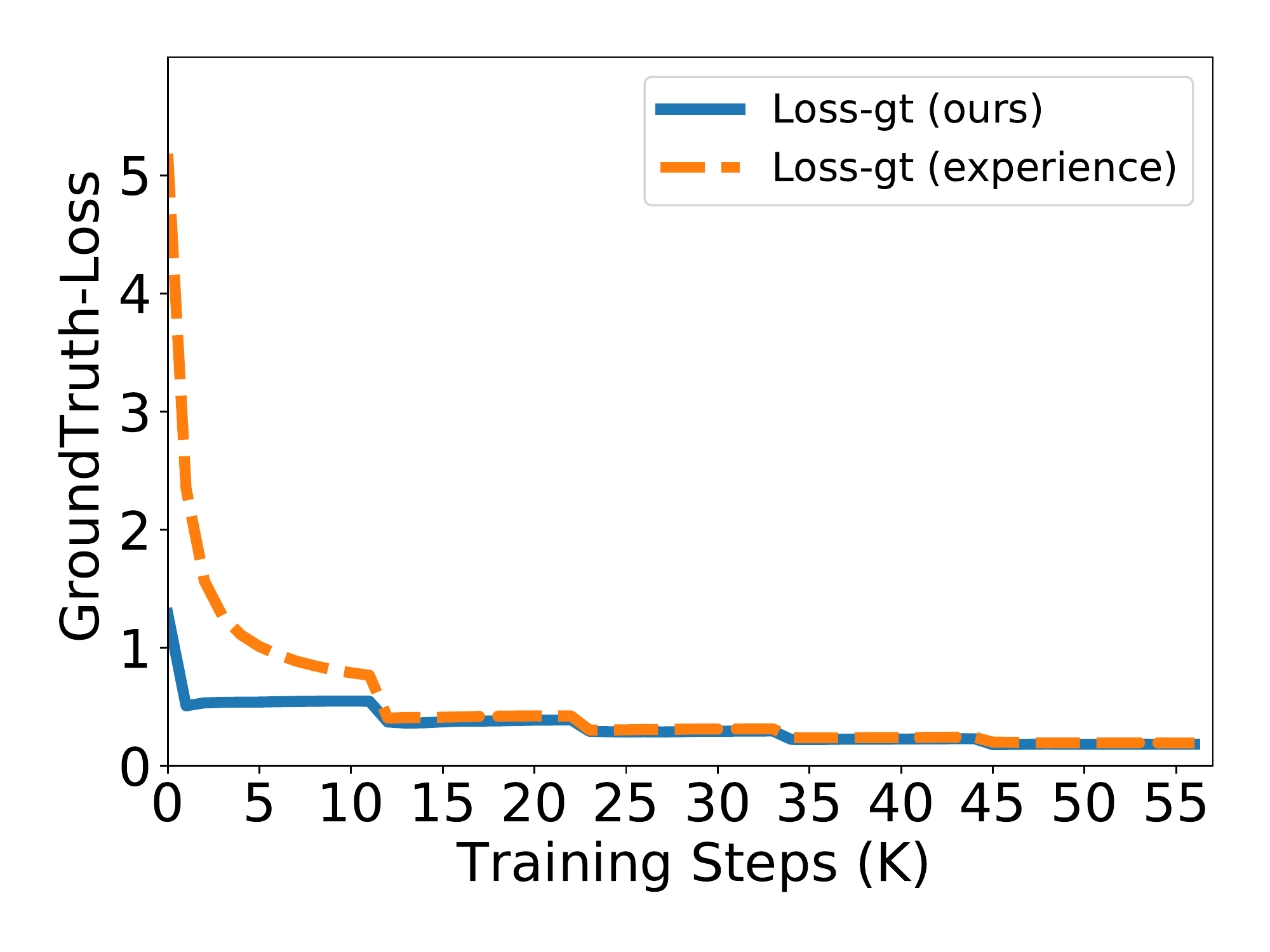}
\end{minipage}
}%
\subfigure[$F1$: 6-bit]{
\begin{minipage}[t]{0.25\linewidth}
\centering
\includegraphics[width=1.45in]{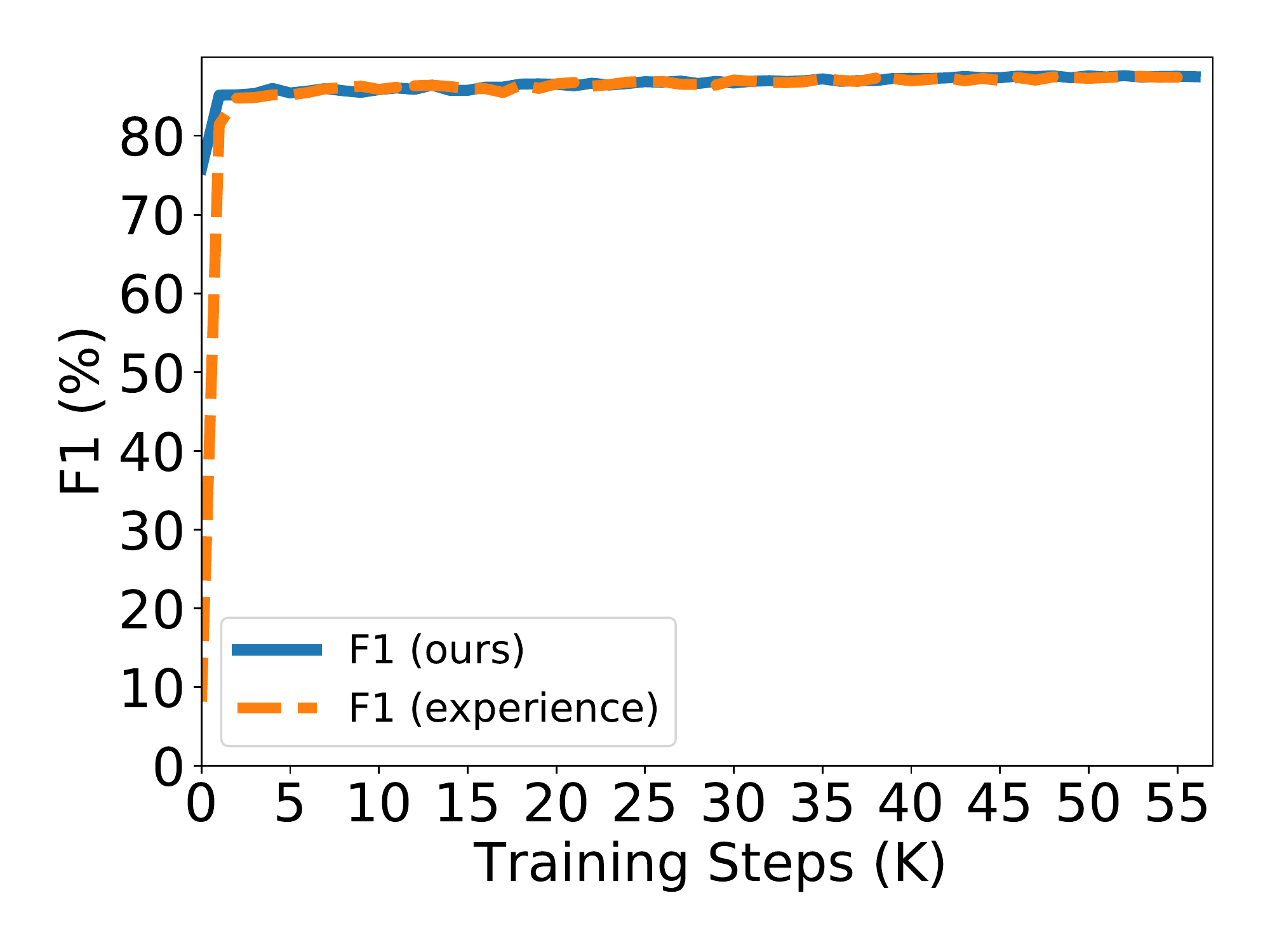}
\end{minipage}
}%
\end{center}
\caption{The impact on \textbf{total training loss},  \textbf{distillation loss}, \textbf{ground truth loss} and \textbf{accuracy}  when implementing our proposed KDLSQ-BERT by using different scale-factor initialization. \textbf{The quantization bit "W-E-A (\#bit)" is set to 6-6-8}. The experimental results are tested by adopting \textbf{"SQuAD 1.1"}  and \textbf{"TinyBERT"}}
\label{figure:9}
\end{figure}

\begin{figure}[htb]
\begin{center}
\subfigure[$Loss_{total}$: 8-bit]{
\begin{minipage}[t]{0.25\linewidth}
\centering
\includegraphics[width=1.45in]{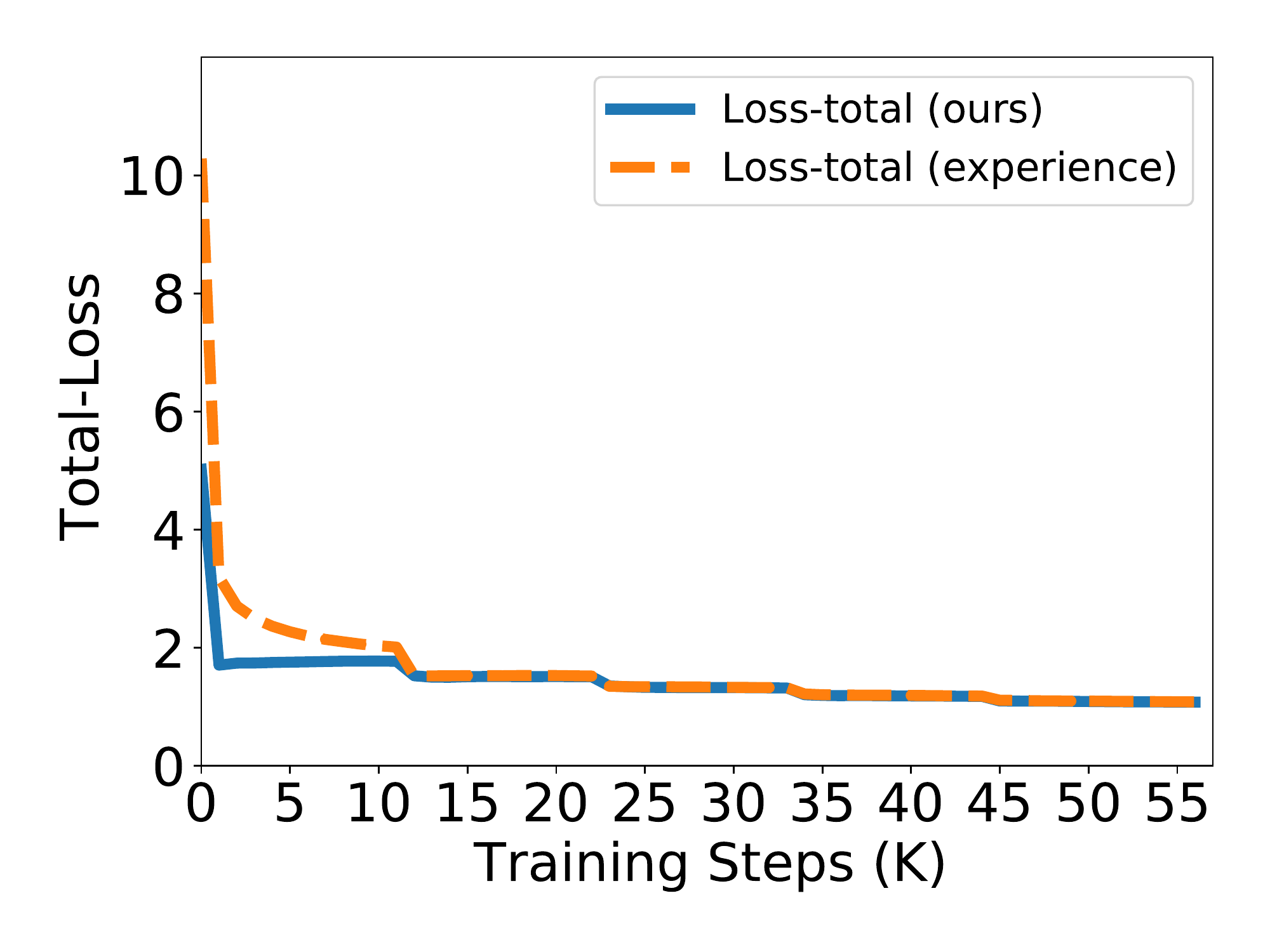}
\end{minipage}%
}%
\subfigure[$Loss_{kd}$: 8-bit]{
\begin{minipage}[t]{0.25\linewidth}
\centering
\includegraphics[width=1.45in]{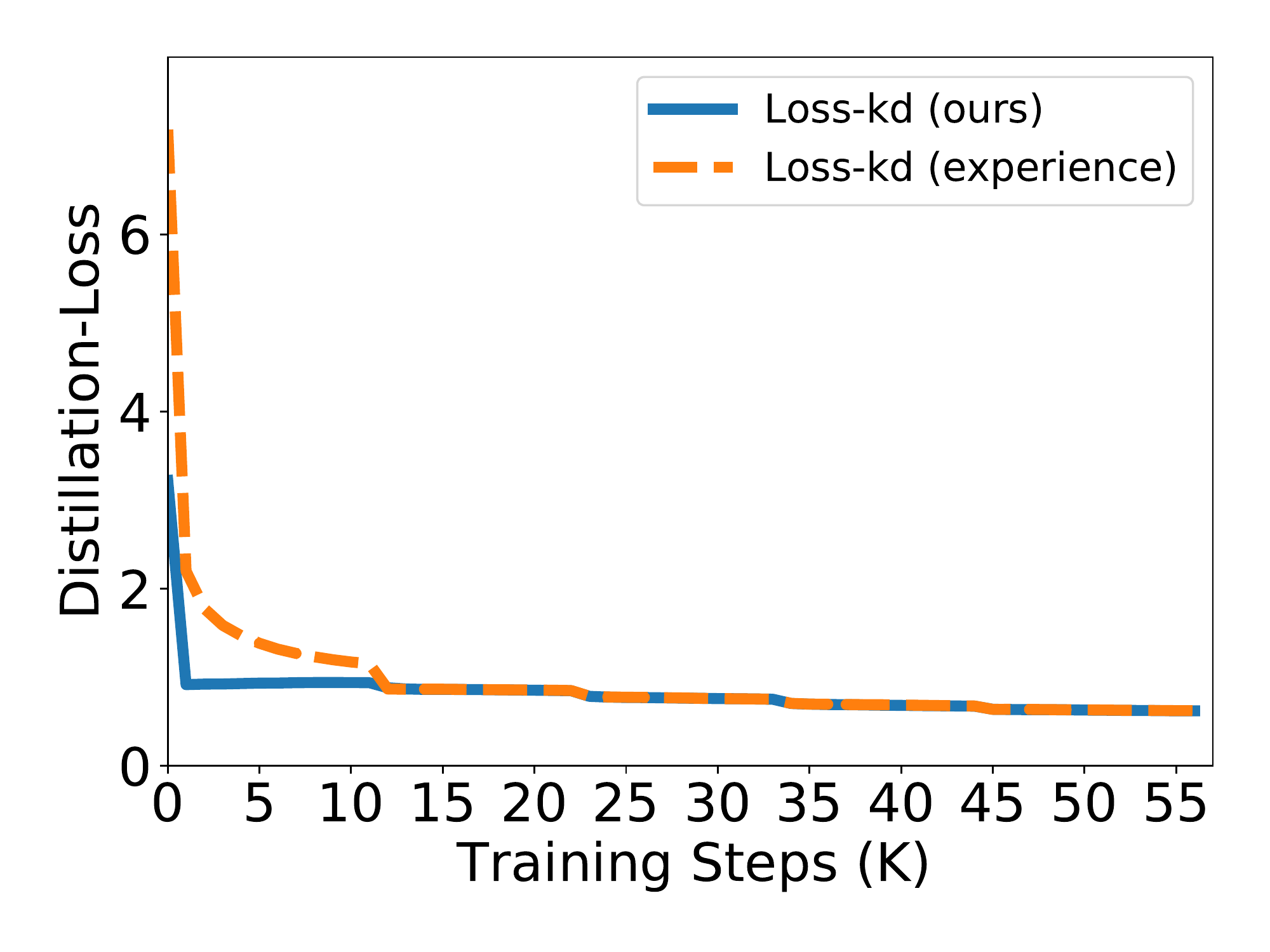}
\end{minipage}%
}%
\subfigure[$Loss_{gt}$: 8-bit]{
\begin{minipage}[t]{0.25\linewidth}
\centering
\includegraphics[width=1.45in]{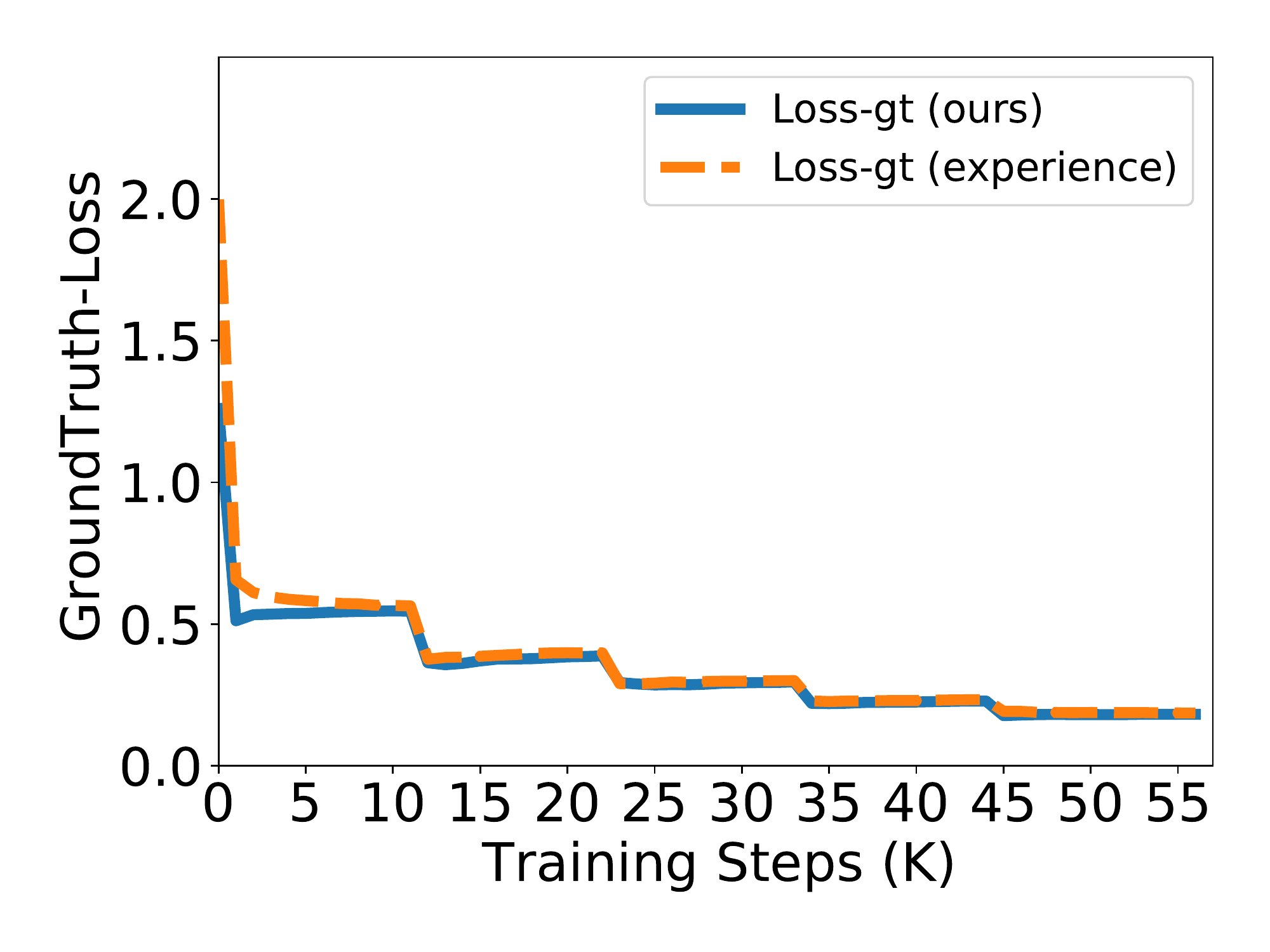}
\end{minipage}
}%
\subfigure[$F1$: 8-bit]{
\begin{minipage}[t]{0.25\linewidth}
\centering
\includegraphics[width=1.45in]{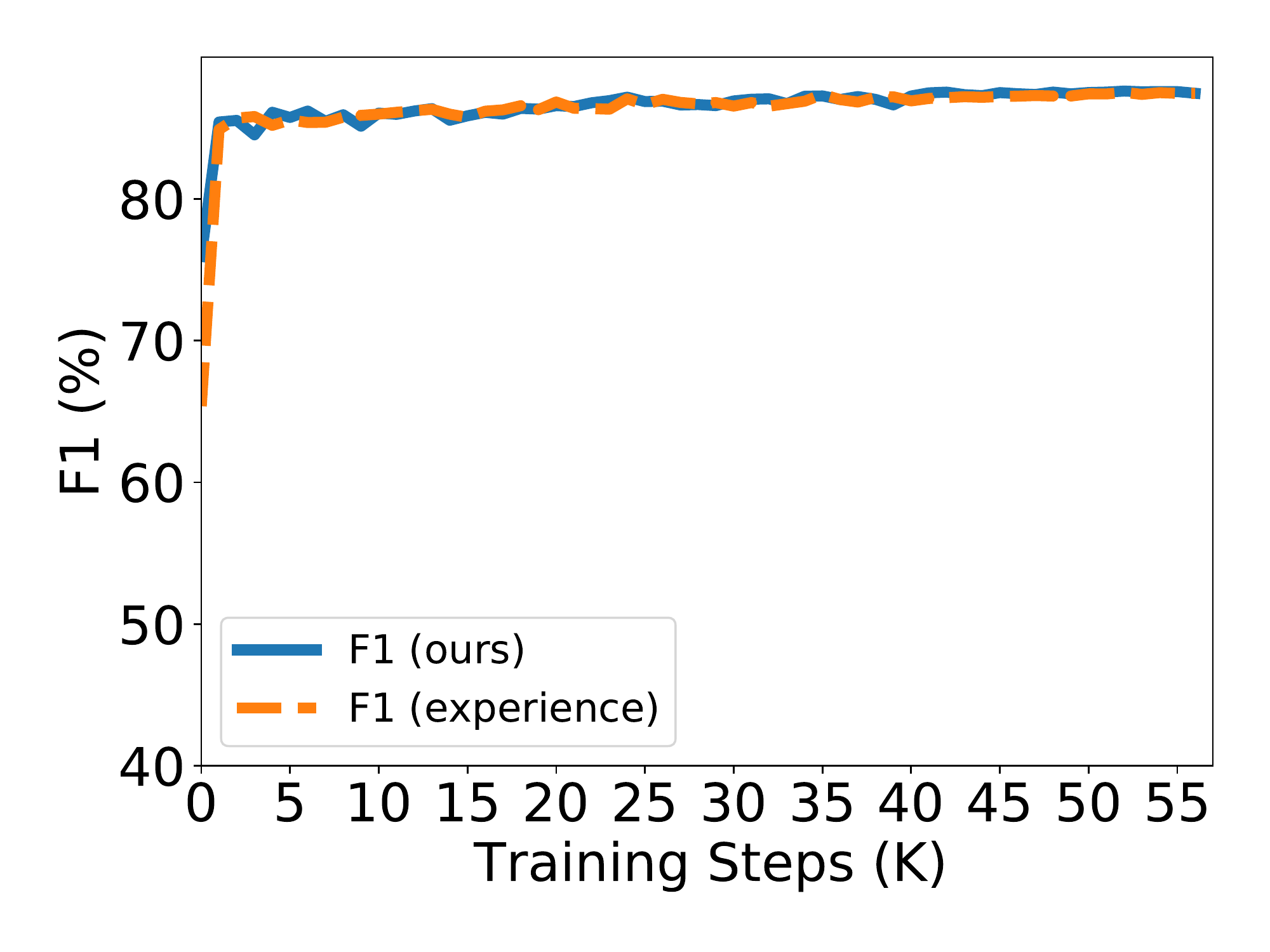}
\end{minipage}
}%
\end{center}
\caption{The impact on \textbf{total training loss},  \textbf{distillation loss}, \textbf{ground truth loss} and \textbf{accuracy}  when implementing our proposed KDLSQ-BERT by using different scale-factor initialization. \textbf{The quantization bit "W-E-A (\#bit)" is set to 8-8-8}. The experimental results are tested by adopting \textbf{"SQuAD 1.1"}  and \textbf{"TinyBERT"}}
\label{figure:10}
\end{figure}

For the "SQuAD 1.1" related to the question-answering task, Figure \ref{figure:7} $\sim$ Figure \ref{figure:10} demonstrate the same results as those of "mnli". As illustrated in those figures, when using Algorithm \ref{alg:scale_factor_initialization} to do initialization for scale-factors, the correlated training losses and model accuracy can get faster convergence speed as the training step goes. This is because the scale-factors can be well initialized by using our scale-factor initialization method, such that an effective truncation for the quantized tensor results in good training convergence. Besides, all correlated training losses can be converged effectively, which proves that it is a good way to enhance the model accuracy by setting both $Loss_{kd}$ and $Loss_{gt}$ as reference for model training.

\subsection{Ablation Studies}
\label{ablation_studies}
According to the steps from Algorithm \ref{alg:distillation_aware_quantization}, our proposed KDLSQ-BERT mainly consists of three components including LSQ, knowledge distillation, as well as the ground truth loss based on training data set. To further analyze the effects of different components, we perform ablation study on different NLP tasks, and the specific comparison results are listed in Table \ref{tab:4} and Table \ref{tab:5}. In these tables, "LSQ+KD+Lgt" denotes our proposed KDLSQ-BERT, "LSQ+KD" indicates LSQ along with knowledge distillation, "LSQ" indicates LSQ method only.

\begin{table}[htb]
\caption{Ablation Study on GLUE benchmark.}
\label{tab:4}
\begin{center}
\scalebox{0.60} {
\begin{tabular}{llcc|ccccccccc}
\hline \hline & & W-E-A & Size &\multirow{2}{*}{MNLI} & \multirow{2}{*}{CoLA} & \multirow{2}{*}{MRPC} & \multirow{2}{*}{QNLI} & \multirow{2}{*}{QQP} & \multirow{2}{*}{RTE} & \multirow{2}{*}{SST-2} & \multirow{2}{*}{STS-B} & \multirow{2}{*}{$Average$}  \\  & & (\#bits) & (MB) & &  & &  &  & & & & \\ \hline
\hline Full-precision & BERT-base  & 32-32-32 & 418 ($\times 1.0$) & 84.463 & 58.081 & 90.625 & 91.964 & 87.762 & 71.119 & 93.119 & 89.823 & 83.369 \\

\hline \multirow{3}{*}{2-bit} & LSQ & 2-2-8 & 28 ($\times 14.9$) & 83.271 & \bf 49.981 & 90.117 & 90.024 & 87.445 & \bf 70.397 & 91.399 & 87.609 & 81.280 \\
& LSQ+KD & 2-2-8 & 28 ($\times 14.9$) & 84.361 & 48.724 & \bf 90.444 & 91.415 & 87.696 & 67.148 & 92.202 & 88.117 & 81.263\\
& LSQ+KD+Lgt & 2-2-8 & 28 ($\times 14.9$) & \bf 84.564 & 49.418 & 90.290 & \bf 91.452 & \bf 88.071 & 67.509 & \bf 92.775 & \bf 88.117 & \bf 81.524 \\
\hline \multirow{3}{*}{4-bit} & LSQ & 4-4-8 & 54 ($\times 7.7$) & 84.218 & 57.738 & 90.183 & 91.305 & 87.827 & 72.924 & 93.005 & 89.569 & 83.359 \\
& LSQ+KD & 4-4-8 & 54 ($\times 7.7$) & 85.064 & 59.274 & 90.694 & \bf 92.239 & 88.027 & \bf 72.924 & \bf 93.693 & 89.697 & 83.951\\
& LSQ+KD+Lgt & 4-4-8 & 54 ($\times 7.7$) & \bf 85.329 & \bf 60.646 & \bf 91.394 & 92.165 & \bf 88.503 & 72.729 & 93.349 & \bf 89.697 & \bf 84.221 \\

\hline \multirow{3}{*}{6-bit} & LSQ & 6-6-8 & 80 ($\times 5.2$) & 84.463 & 60.749 & 85.583 & 91.781 & 88.042 & \bf 74.368 & 93.119 & 89.495 &  83.950\\
& LSQ+KD & 6-6-8 & 80 ($\times 5.2$) & 85.257 & 59.426 & \bf 91.312 & 92.165 & 87.987 & 71.841 & 93.463 & 90.000 & 83.931\\
& LSQ+KD+Lgt & 6-6-8 & 80 ($\times 5.2$) & \bf 85.329 & \bf 61.482 & 91.003 & \bf 92.367 & \bf 88.463 & 71.480 & \bf 93.807 & \bf 90.000 & \bf 84.241 \\

\hline \multirow{3}{*}{8-bit} & LSQ & 8-8-8 & 106 ($\times 3.9$) & 84.401 & 59.937 & 90.941 & 91.122 & 87.832 & \bf 72.202 & 93.349 & 89.569 & 83.669 \\
& LSQ+KD & 8-8-8 & 106 ($\times 3.9$) & 85.084 & 59.994 & 91.096 & \bf 92.165 & 87.947 & 71.841 & \bf 93.807 & 89.944 & 83.985\\
& LSQ+KD+Lgt & 8-8-8 & 106 ($\times 3.9$) & \bf 85.308 & \bf 60.774 & \bf 91.826 & 92.147 & \bf 88.439 & 71.841 & 93.693 & \bf 89.944 & \bf 84.247 \\
\hline
\hline  Full-precision & TinyBERT  & 32-32-32 & 258 ($\times 1.6$)  & 84.768 & 54.175 & 91.319 & 90.793 & 87.966 & 71.841 & 90.252 & 89.792 & 82.613\\
\hline \multirow{3}{*}{2-bit} & LSQ & 2-2-8 & 18 ($\times 23.2$) & 83.179 & 49.344 & 90.501 & 88.285 & 87.635 &	63.899 & 91.514 & 87.704 & 80.257 \\
& LSQ+KD & 2-2-8 & 18 ($\times 23.2$) &  \bf 83.953 & \bf 49.498 & 90.690 & 90.170 & 87.642 & 62.455 & 92.661 & 87.539 & 80.576\\
& LSQ+KD+Lgt & 2-2-8 & 18 ($\times 23.2$) & 83.902 &	49.047 & \bf 91.100 & \bf 90.518 & \bf 88.038 & \bf 64.260 &  \bf 92.661 &  \bf 87.539 & \bf 80.883 \\

\hline \multirow{3}{*}{4-bit} & LSQ  & 4-4-8 & 34 ($\times 12.3$) & 84.340 & 52.588 & 89.949 & 90.170 & 87.944 &	\bf 74.368 & 92.431 & 89.767 & 82.695 \\
& LSQ+KD & 4-4-8 & 34 ($\times 12.3$) & 84.727 &	54.062 & 91.161 & 91.269 & 87.984 & 71.119 & 92.775 & 89.825 & 82.865 \\
& LSQ+KD+Lgt & 4-4-8 & 34 ($\times 12.3$) & \bf 84.962 &	\bf 54.415 & \bf 91.192 & \bf 91.671 & \bf 88.300 & 72.202 & \bf 92.890 & \bf 89.825 & \bf 83.182 \\

\hline \multirow{3}{*}{6-bit} & LSQ & 6-6-8 & 50 ($\times 8.3$) & 84.432 & 53.321 & 89.865 & 90.591 & 88.030 &	\bf 76.173 & 92.202 & 88.041 & 82.832  \\
& LSQ+KD & 6-6-8 & 50 ($\times 8.3$) & 84.819 &	\bf 54.302 & \bf 91.478 & 91.177 & 88.007 &	71.841 &93.006 & 89.895 & 83.066 \\
& LSQ+KD+Lgt & 6-6-8 & 50 ($\times 8.3$) & \bf 85.013 &	54.293 & 91.035& \bf 91.580 & \bf 88.333 & 73.285 & \bf 93.006 & \bf 89.895 & \bf 83.305\\
\hline \multirow{3}{*}{8-bit} & LSQ  & 8-8-8 & 65 ($\times 6.4$) & 84.615 &	53.807 & 90.508 & 90.957 & 88.092 & \bf 76.173 & 92.288 & 89.950 & 83.299 \\
& LSQ+KD & 8-8-8 & 65 ($\times 6.4$) & 84.911 &	54.521 & \bf 91.319 & 91.140 & 88.039 &	72.563 & 92.775 & 89.958 & 83.153 \\
& LSQ+KD+Lgt & 8-8-8 & 65 ($\times 6.4$) & \bf 85.013 & \bf 54.793 & 91.130 & \bf 91.580 & \bf 88.315& 73.646 & \bf 92.890 & \bf 89.958 & \bf 83.416\\
\hline
\hline
\end{tabular}}
\end{center}
\end{table}

\begin{table}[htb]
\caption{Ablation Study on SQuAD.}
\label{tab:5}
\begin{center}
\scalebox{0.68} {
\begin{tabular}{llcc|ccc}
\hline \hline & & W-E-A & Size &\multirow{2}{*}{SQuAD 1.1}  & \multirow{2}{*}{SQuAD 2.0} & \multirow{2}{*}{$Average$}\\  &  & (\#bits) & (MB) & & & \\ \hline
\hline Full-precision & BERT-base & 32-32-32 & 418 ($\times 1.0$) & 88.696 & 77.725 & 83.210\\ \hline
\multirow{3}{*}{2-bit} & LSQ & 2-2-8 & 28 ($\times 14.9$) & 84.105 & 74.254 & 79.179\\
& LSQ+KD & 2-2-8 & 28 ($\times 14.9$) & \bf 88.794 & 77.619 & 83.207 \\
& LSQ+KD+Lgt & 2-2-8 & 28 ($\times 14.9$) & 88.447 & \bf 78.400 & \bf 83.423 \\
\hline
\multirow{3}{*}{4-bit} & LSQ & 4-4-8 & 54 ($\times 7.7$) & 86.098 & 76.840 & 81.469\\
& LSQ+KD & 4-4-8 & 54 ($\times 7.7$) & \bf 89.236 & 78.421 & 83.828 \\
& LSQ+KD+Lgt & 4-4-8 & 54 ($\times 7.7$) & 89.207 & \bf 78.965 & \bf 84.086 \\
\hline
\multirow{3}{*}{6-bit} & LSQ & 6-6-8 & 80 ($\times 5.2$) & 86.357 & 76.438 & 81.398\\
& LSQ+KD & 6-6-8 & 80 ($\times 5.2$) & \bf 89.296 & 78.412 & 83.854 \\
& LSQ+KD+Lgt & 6-6-8 & 80 ($\times 5.2$) & 89.280 & \bf 78.905 & \bf 84.092 \\
\hline
\multirow{3}{*}{8-bit} & LSQ & 8-8-8 & 106 ($\times 3.9$) & 86.576 & 76.334  & 81.455\\
& LSQ+KD & 8-8-8 & 106 ($\times 3.9$) & \bf 89.292 & 78.500 & 83.896 \\
& LSQ+KD+Lgt & 8-8-8 & 106 ($\times 3.9$) & 89.218 & \bf 79.330 & \bf 84.274 \\
\hline
\hline Full-precision & TinyBERT  & 32-32-32 & 258 ($\times 1.6$) & 87.527 & 77.730 & 82.629\\ \hline
\multirow{3}{*}{2-bit} & LSQ & 2-2-8 & 18 ($\times 23.2$) & 84.769 & 73.518 & 79.144\\
& LSQ+KD & 2-2-8 & 18 ($\times 23.2$) & 86.053 &	76.747 & 81.400 \\
& LSQ+KD+Lgt & 2-2-8 & 18 ($\times 23.2$) & \bf 86.635 &	\bf 76.889 & \bf 81.762 \\
\hline
\multirow{3}{*}{4-bit} & LSQ & 4-4-8 & 34 ($\times 12.3$) & 85.667 & 75.304 & 80.486\\
& LSQ+KD & 4-4-8 & 34 ($\times 12.3$) & 87.495 &	77.967 & 82.731 \\
& LSQ+KD+Lgt & 4-4-8 & 34 ($\times 12.3$) & \bf 87.887 & \bf 77.981 & \bf 82.934 \\
\hline
\multirow{3}{*}{6-bit} & LSQ & 6-6-8 & 50 ($\times 8.3$) & 85.902 & 75.640 & 80.771\\
& LSQ+KD & 6-6-8 & 50 ($\times 8.3$) & 87.581 &  78.067 & 82.824 \\
& LSQ+KD+Lgt & 6-6-8 & 50 ($\times 8.3$) & \bf 87.944 & \bf 78.279 & \bf 83.111 \\
\hline
\multirow{3}{*}{8-bit} & LSQ & 8-8-8 & 65 ($\times 6.4$) & 86.587 & 75.908 & 81.247\\
& LSQ+KD & 8-8-8 & 65 ($\times 6.4$) & 87.587 & 78.079 & 82.833 \\
& LSQ+KD+Lgt & 8-8-8 & 65 ($\times 6.4$) & \bf 87.948 & \bf 78.302 & \bf 83.125 \\
\hline
\hline
\end{tabular}}
\end{center}
\end{table}

From the tables, it demonstrates that for both the GLUE benchmark and the SQuAD, "LSQ+KD+Lgt" can perform better accuracy than both "LSQ" and "LSQ+KD" on average. This is because the training loss for "LSQ+KD+Lgt" involves more reference information to improve the training result. The correlated information mainly include knowledge from both the teacher model and the ground truth labels of the training data. Different from "LSQ+KD+Lgt", "LSQ" involves only knowledge from the ground truth labels of the training data, and "LSQ+KD" involves only the knowledge of the teacher model. Note that the experimental results show that for the GLUE benchmark, "LSQ" and "LSQ+KD" can obtain almost the same accuracy in the same bit quantization; while for the SQuAD, "LSQ+KD" can perform about 2.0\% higher accuracy than "LSQ" when doing the same bit quantization. As a result, it is hard to determine if "LSQ+KD" is better than "LSQ" or not because both methods apply different knowledge for quantization training. However, the results from the tables prove that KDLSQ-BERT (e.g., "LSQ+KD+Lgt") performs better not only in high-bit (e.g., 8-bit) quantization, but also in low-bit (e.g., 2-bit) quantization. Consequently, KDLSQ-BERT is an effective and practical training method for BERT quantization.

\subsection{Experimental Analysis on Low Bit Quantization}
\label{low_bit_quantization}
To further study the performance of our method on different NLP tasks, we conduct KDLSQ-BERT by setting KDLSQ-BERT in ultra-low bit quantization. The specific experimental results are presented in Table \ref{tab:6} and Table \ref{tab:7}, respectively. As shown in the tables, it demonstrates that for both the GLUE benchmark and the SQuAD, our quantization method can obtain the same accuracy as the full-precision base-line model even if doing the ultra-low bit quantization. Especially, it should be noted that compared with the quantization configuration of "4-4-2" and "2-2-2", setting "2-2-4" can get almost the same accuracy performance as the full-precision base-line model. On the one hand, it proves that activation quantization is more sensitive to affect the accuracy performance than the weight quantization. On the other hand, these empirical results tell us that it 2-bit is enough for weight quantization when using our method to do quantization training. Therefore, even if conducting ultra-low bit quantization, our method not only can guarantee the accuracy performance, but also can get an impressive compression ratio.

\begin{table}[htb]
\caption{Low bit quantization of KDLSQ-BERT on the GLUE benchmark.}
\label{tab:6}
\begin{center}
\scalebox{0.60} {
\begin{tabular}{llcc|ccccccccc}
\hline \hline & & W-E-A & Size &\multirow{2}{*}{MNLI} & \multirow{2}{*}{CoLA} & \multirow{2}{*}{MRPC} & \multirow{2}{*}{QNLI} & \multirow{2}{*}{QQP} & \multirow{2}{*}{RTE} & \multirow{2}{*}{SST-2} & \multirow{2}{*}{STS-B} & \multirow{2}{*}{$Average$}  \\  & & (\#bit) & (MB) & &  & &  &  & & & & \\ \hline
\hline Full-precision & BERT-base  & 32-32-32 & 418 ($\times 1.0$) & 84.463 & 58.081 & 90.625 & 91.964 & 87.762  & 71.119 & 93.119 	& 89.823  & 83.369 \\
\hline \multirow{1}{*}{2-bit} & KDLSQ-BERT (ours) & 2-2-2 & 28 ($\times 14.9$) & 83.607 & 54.885 & 90.000 & 90.536 & 87.962 & 66.065 & 92.775 & 85.854 & 81.460 \\
\hline \multirow{1}{*}{2-bit} & KDLSQ-BERT (ours) & 2-2-4 & 28 ($\times 14.9$) & 85.013 & 58.081 & 91.388 & 92.092 & 87.996 & 73.646 & 93.693  & 89.989  & 83.987\\
\hline \multirow{1}{*}{4-bit} & KDLSQ-BERT (ours) & 4-4-2 & 54 ($\times 7.7$) & 83.535 & 54.844 & 89.619 & 90.902  & 88.085 & 67.870 	&  93.005 & 87.095 & 81.869\\
\hline \multirow{1}{*}{4-bit} & KDLSQ-BERT (ours) & 4-4-4 & 54 ($\times 7.7$) &  \bf 85.115  & 60.132  & 92.063 & 92.257  & 88.355  & \bf 73.285 & 93.234 & 89.851  & 84.287 \\
\hline \multirow{1}{*}{4-bit} & KDLSQ-BERT (ours) & 4-4-6 & 54 ($\times 7.7$) & 85.094   & 	\bf 61.788  & \bf 92.171   & \bf 92.367   & \bf 88.526 	& 71.480  & \bf 93.693  & \bf 90.197   & \bf 84.414 \\
\hline \multirow{1}{*}{4-bit} & KDLSQ-BERT (ours) & 4-4-8 & 54 ($\times 7.7$) & 84.218 & 57.738 & 90.183 & 91.305 & 87.827 & 72.924 & 93.005 & 89.569 & 83.359 \\
\hline
\hline  Full-precision & TinyBERT  & 32-32-32 & 258 ($\times 1.6$)  & 84.768 & 54.175 & 91.319 & 90.793 & 87.966 & 71.841 & 90.252 & 89.792 & 82.613 \\
\hline \multirow{1}{*}{2-bit} & KDLSQ-BERT (ours) & 2-2-2 & 18 ($\times 23.2$) & 82.354 & 45.805 & 89.003 & 89.694 & 87.721 & 65.343 & 92.431 & 84.460 & 79.602 \\
\hline \multirow{1}{*}{2-bit} & KDLSQ-BERT (ours) & 2-2-4 & 18 ($\times 23.2$) & 84.483 & 54.944 & 90.909 & 91.177 & 88.376 & 75.451 & 93.119 & 89.420 & 83.485 \\
\hline \multirow{1}{*}{4-bit} & KDLSQ-BERT (ours) & 4-4-2 & 34 ($\times 12.3$) & 82.517 & 47.306 & 89.608 & 89.658 & 88.005 & 65.704 & 92.202 & 85.430 & 80.054 \\
\hline \multirow{1}{*}{4-bit} & KDLSQ-BERT (ours) & 4-4-4 & 34 ($\times 12.3$) & 84.697 & 55.130 & 90.630 & 91.580 & \bf 88.414 & 74.007 & 93.119 & 89.580 & 83.395 \\
\hline \multirow{1}{*}{4-bit} & KDLSQ-BERT (ours) & 4-4-6 & 34 ($\times 12.3$) & 84.768 & \bf 55.727 & \bf 91.388 & \bf 91.781 & 88.394 & \bf 75.812 & \bf 93.234 & \bf 89.952 & \bf 83.882 \\
\hline \multirow{1}{*}{4-bit} & KDLSQ-BERT (ours) & 4-4-8 & 34 ($\times 12.3$) & \bf 84.962 & 54.415 & 91.192 & 91.671 & 88.300 & 72.202 	& 92.890 & 89.825 & 83.182 \\
\hline
\hline
\end{tabular}}
\end{center}
\end{table}

\begin{table}[htb]
\caption{Low bit quantization of KDLSQ-BERT on the SQuAD.}
\label{tab:7}
\begin{center}
\scalebox{0.68} {
\begin{tabular}{llcc|ccc}
\hline \hline & & W-E-A & Size &\multirow{2}{*}{SQuAD 1.1}  & \multirow{2}{*}{SQuAD 2.0} & \multirow{2}{*}{$Average$}\\  &  & (\#bit) & (MB) & & & \\ \hline
\hline Full-precision & BERT-base & 32-32-32 & 418 ($\times 1.0$) & 88.696 & 77.725 & 83.210\\ \hline
\multirow{1}{*}{2-bit} & KDLSQ-BERT (ours) & 2-2-2 & 28 ($\times 14.9$) & 84.818 & 73.339 & 79.078\\
\hline
\multirow{1}{*}{2-bit} & KDLSQ-BERT (ours) & 2-2-4 & 28 ($\times 14.9$) & 88.996 & 78.213 & 83.604\\
\hline
\multirow{1}{*}{4-bit} & KDLSQ-BERT (ours) & 4-4-2 & 54 ($\times 7.7$) & 85.070 & 77.725 & 81.397\\
\hline
\multirow{1}{*}{4-bit} & KDLSQ-BERT (ours) & 4-4-4 & 54 ($\times 7.7$) & 88.998 & 77.919 & 83.458\\
\hline
\multirow{1}{*}{4-bit} & KDLSQ-BERT (ours) & 4-4-6 & 54 ($\times 7.7$) & 89.070 & 78.342 & 83.706\\
\hline
\multirow{1}{*}{4-bit} & KDLSQ-BERT (ours) & 4-4-8 & 54 ($\times 7.7$) & \bf 89.207 & \bf 78.965 & \bf 84.086\\
\hline
\hline Full-precision & TinyBERT  & 32-32-32 & 258 ($\times 1.6$) & 87.527 & 77.730 & 82.629\\ \hline
\multirow{1}{*}{2-bit} & KDLSQ-BERT (ours) & 2-2-2 & 18 ($\times 23.2$) & 80.695 & 70.124 & 75.410\\
\hline
\multirow{1}{*}{2-bit} & KDLSQ-BERT (ours) & 2-2-4 & 18 ($\times 23.2$) & 86.858 & 77.197 & 82.027\\
\hline
\multirow{1}{*}{4-bit} & KDLSQ-BERT (ours) & 4-4-2 & 34 ($\times 12.3$) & 81.099 & 70.794 & 75.946\\
\hline
\multirow{1}{*}{4-bit} & KDLSQ-BERT (ours) & 4-4-4 & 34 ($\times 12.3$) & 86.902 & 77.400 & 82.151\\
\hline
\multirow{1}{*}{4-bit} & KDLSQ-BERT (ours) & 4-4-6 & 34 ($\times 12.3$) & 87.569 & 77.796 & 82.682\\
\hline
\multirow{1}{*}{4-bit} & KDLSQ-BERT (ours) & 4-4-8 & 34 ($\times 12.3$) & \bf 87.887 & \bf 77.981 & \bf 82.934\\
\hline
\hline
\end{tabular}}
\end{center}
\end{table}

\section{Conclusion and Future Work}
\label{Conclusion and Future Work}
In this work, we proposed a novel quantization method KDLSQ-BERT to quantize the Transformer-based models such as BERT. In addition to exploiting the ground truth for training loss calculation, the main feature of KDLSQ-BERT is that the KD technique is leveraged to transfer the knowledge from a "teacher" BERT to a "student" BERT when doing LSQ to quantize that "student" BERT. Empirical experiments show that KDLSQ-BERT not only can outperform the state-of-the-art BERT quantization methods, but also can achieve high accuracy when doing ultra low-bit (e.g., 2-bit) weight quantization, and even can perform comparable performance as the full-precision base-line model. In the near future we will explore how to quantize other NLP models in low-bit (such as 2-bit) on the basis of LSQ.

\bibliography{iclr2021_conference}
\bibliographystyle{iclr2021_conference}

\end{document}